\documentclass[iicol,sn-basic]{sn-jnl}% Math and Physical Sciences Numbered Reference Style

\usepackage{graphicx}
\usepackage{multirow}
\usepackage{amsmath,amssymb,amsfonts}
\usepackage{amsthm}
\usepackage{mathrsfs}
\usepackage[title]{appendix}
\usepackage{textcomp}
\usepackage{manyfoot}
\usepackage{booktabs}
\usepackage{array}
\usepackage{algorithm}
\usepackage{algorithmicx}
\usepackage{algpseudocode}

\usepackage{listings}
\usepackage[table,xcdraw]{xcolor}
\usepackage{colortbl}
\usepackage{hhline}
\usepackage{arydshln}
\usepackage{adjustbox}
\usepackage{wrapfig}
\usepackage[normalem]{ulem}
\usepackage{tikz}
\usepackage{comment}
\usepackage[utf8]{inputenc}
\usepackage[T1]{fontenc}
\usepackage{hyperref}
\usepackage{url}
\usepackage{nicefrac}
\usepackage{microtype}
\usepackage{lipsum}
\usepackage{footmisc}
\usepackage{makecell}
\usepackage{tabularx}
\definecolor{darkgreen}{rgb}{0.0, 0.5, 0.0}
\definecolor{lightblue}{rgb}{0.93,0.95,1.0}
\definecolor{lightshade}{rgb}{0.9,0.9,0.9}

\theoremstyle{thmstyleone}%
\theoremstyle{thmstyletwo}%

\theoremstyle{thmstylethree}%

\begin{document}

\title[Article Title]{Region Matters: Efficient and Reliable Region-Aware \\ Visual Place Recognition}

\author[1]{Shunpeng Chen}
\author[1]{Yukun Song}
\author[2]{Changwei Wang}
\author[3]{Rongtao Xu}
\author[2]{Kexue Fu}
\author[2]{Longxiang Gao}
\author[1]{Li Guo}
\author[4]{Ruisheng Wang}
\author*[1]{Shibiao Xu}\email{shibiaoxu@bupt.edu.cn}

\affil[1]{School of Artificial Intelligence, Beijing University of Posts and Telecommunications, Beijing 100876, China}
\affil[2]{Key Laboratory of Computing Power Network and Information Security, Ministry of Education, Shandong Computer Science Center, Qilu University of Technology}
\affil[3]{Spatialtemporal AI}
\affil[4]{School of Architecture and Urban Planning, Guangdong-Hong Kong-Macau Joint Laboratory for Smart Cities, Shenzhen University, Shenzhen 518060, China}


\abstract{Visual Place Recognition (VPR) determines a query image's geographic location by matching it against geotagged databases. However, existing methods struggle with perceptual aliasing caused by irrelevant regions and inefficient re-ranking due to rigid candidate scheduling. To address these issues, we introduce FoL++, a method combining robust discriminative region modeling with adaptive re-ranking. Specifically, we propose a Reliability Estimation Branch to generate spatial reliability maps that explicitly model occlusion resistance. This representation is further optimized by two spatial alignment losses (SAL and SCEL) to effectively align features and highlight salient regions. For weakly supervised learning without manual annotations, a pseudo-correspondence strategy generates dense local feature supervision directly from aggregation clusters. Our Adaptive Candidate Scheduler dynamically resizes candidate pools based on global similarity. By weighting local matches by reliability and adaptively fusing global and local evidence, FoL++ surpasses traditional independent matching systems. Extensive experiments across seven benchmarks demonstrate that FoL++ achieves state-of-the-art performance with a lightweight memory footprint, improving inference speed by 40\% over FoL. Code and models will be released (and merged with FoL) at \small{\url{https://github.com/chenshunpeng/FoL}}.}


\maketitle

\section{Introduction}
\label{sec:introduction}

\begin{figure*}[t]
    \centering
    \includegraphics[width=\linewidth]{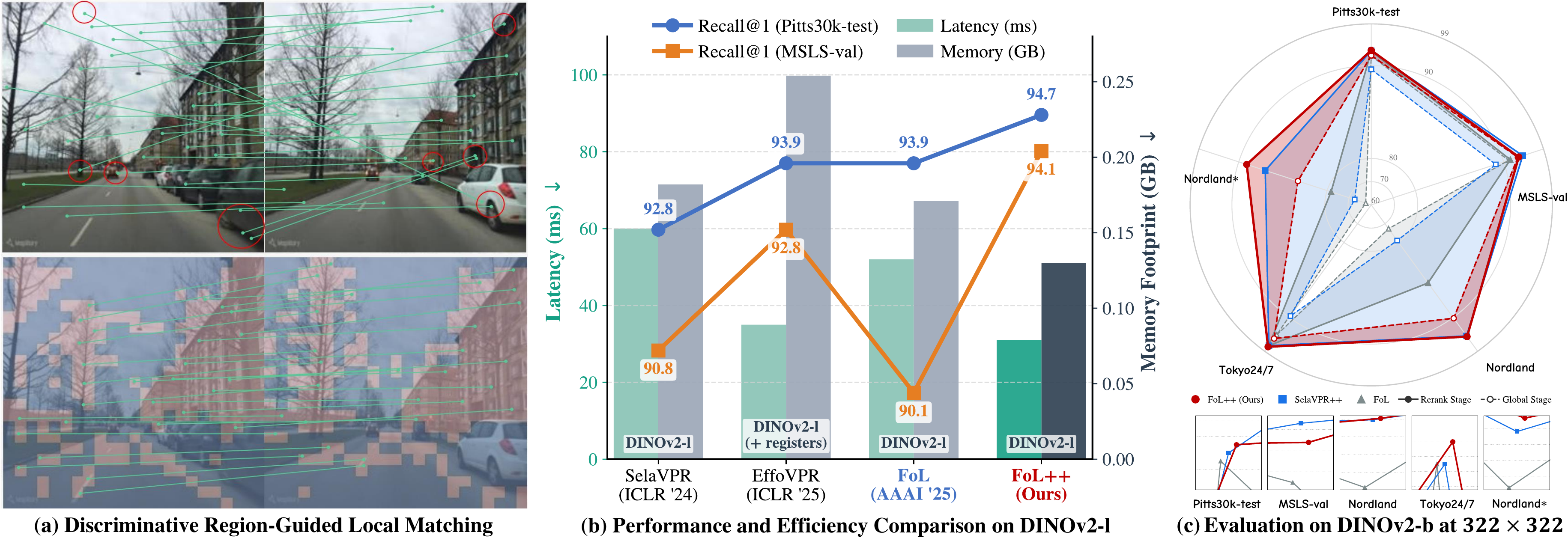}
    \vspace{-0.1cm}
    \caption{Comprehensive evaluation of the proposed FoL++ framework. (a) Visual comparison between standard matching (top) and our discriminative region-guided matching (bottom), showing the effective suppression of mismatches in irrelevant backgrounds. (b) Trade-off analysis demonstrating that FoL++ achieves optimal performance across all metrics. All methods are evaluated on an NVIDIA A100 GPU at $322 \times 322$ resolution, except EffoVPR ($504 \times 504$). (c) Radar chart illustrating that our FoL++ consistently exhibits the strongest overall performance in both the global and re-ranking stages.}
    \label{fig:intro}
    \vspace{-0.2cm}
\end{figure*}

Visual Place Recognition (VPR), also known as geo-localization~ \citep{selavpr,liu2024npr,ali2023global,cosplace}, is essential for various location-aware applications such as augmented reality~ \citep{middelberg2014scalable}, SLAM~ \citep{zou2012coslam,zaffar2021vpr}, visual localization~ \citep{wenzel20254seasons,xu2024local}, and autonomous driving~ \citep{zou2025mim4d,mao20233d}. Fundamentally, VPR addresses the question, “Where was this image captured?” by enabling machines to infer geographic locations from visual cues. At its core, this involves determining whether two images depict the same scene and precisely identifying matching key parts. Ideally, this process should mimic human intuition: we first identify sufficiently discriminative features in one image and search for corresponding ones in another, naturally ignoring non-salient features such as moving cars or changing foliage, while remaining unaffected by various noises or environmental conditions. This innate human ability raises an intriguing question: In the absence of depth or pose information, can neural networks learn robust key feature extraction from very limited supervision signals to indicate whether two images depict the same scene? However, realizing this in real-world scenarios is challenging. Street-level images often include dynamic elements like pedestrians and vehicles that obscure key landmarks~ \citep{msls}. In addition, repetitive architectural patterns and drastic changes in viewpoint, introduce significant perceptual ambiguities~ \citep{cosplace}. These challenges are compounded by illumination variations from day to night and seasonal changes~ \citep{barbarani2023local}, highlighting the urgent need for robust feature representations that remain discriminative under diverse conditions.

Earlier VPR methods leveraged handcrafted descriptors like SIFT \citep{SIFT}, SURF \citep{SURF}, and ORB~ \citep{rublee2011orb} to extract discriminative features. These features were aggregated via techniques such as BoW \citep{BoW}, FV \citep{jegou2010aggregating}, GeM \citep{gem}, and NetVLAD \citep{arandjelovic2016netvlad}, forming compact descriptors for retrieval-based localization.
While traditional contrastive learning methods typically rely on hard sample mining, recent advances demonstrate that employing graded similarity supervision derived from camera pose priors can obviate the need for such mining strategies, improving data efficiency and training speed \citep{leyva2023data}. Another alternative to bypass explicit pairwise comparisons is to frame VPR as a classification problem \citep{cosplace,eigenplaces,Divide_classify}. In this approach, place representations are learned by dividing the environment into spatial grids based on geographic coordinates, where each grid cell is assigned a unique class label, though its effectiveness hinges on large-scale training data.

Recent advancements have adopted Visual Foundation Models (VFMs), such as DINOv2~ \citep{dinov2}, for generating global image representations. By aggregating local features, these models yield semantically rich global descriptors~ \citep{anyloc,cricavpr}. However, relying solely on global features often neglects spatial geometric information, rendering the system susceptible to perceptual aliasing, particularly in challenging long-term and dynamic localization scenarios~ \citep{boq}. To address this limitation, the two-stage VPR paradigm~ \citep{patchvlad, transvpr, r2former} has gained significant traction. This approach first retrieves the top-$k$ candidates from the database via global retrieval and subsequently re-ranks them using local feature matching, achieving performance improvements.

While this hierarchical pipeline of ``global retrieval followed by local verification''~ \citep{patchvlad,sferrazza2025match} effectively boosts recall rates, traditional geometric verification steps remain computationally prohibitive for large-scale applications~ \citep{noh2017large,transvpr}. 
Moreover, existing re-ranking strategies, such as EffoVPR~\citep{effovpr} and SelaVPR~\citep{selavpr}, suffer from inefficiencies. Although they utilize multi-layer dense features or upsampled descriptors to ensure broad spatial coverage, they typically perform exhaustive matching across the entire representation without explicitly filtering out background noise. This lack of discriminative region selection inevitably wastes computational resources on irrelevant areas (see Fig.~\ref{fig:intro}(a) and (b)). Furthermore, existing frameworks treat global similarity assessment and local verification as disconnected components~ \citep{patchvlad,r2former,effovpr}, leading to suboptimal candidate ranking. Based on this analysis, we identify three core challenges inherent in current approaches:
\begin{enumerate}
    \item \textbf{Inadequate Spatial Reliability Modeling:} Traditional methods often lack explicit mechanisms to identify spatially stable regions against extreme viewpoint changes and severe occlusions. Although conceptually distinct, these challenges frequently compound, where severe viewpoint shifts naturally induce foreground occlusions. Failing to distinguish stable structures from transient occluders causes severe perceptual aliasing, as confirmed in Fig.~\ref{fig:intro}(a).
    \item \textbf{Limited Supervision Signals:} The lack of correspondence annotations in standard VPR datasets hinders the development of robust local feature representations. Without fine-grained supervision, local descriptors fail to generalize effectively to real-world environments.
    \item \textbf{Static Scheduling Mechanisms:} Existing computational pipelines rely on fixed thresholds for candidate selection and re-ranking. This static approach lacks the flexibility to adapt to dynamic variations in scene complexity, resulting in inefficient resource allocation.
\end{enumerate}

Despite the leading performance of our previous work, FoL~ \citep{FoL}\footnote{Open-sourced at \href{https://github.com/chenshunpeng/FoL}{https://github.com/chenshunpeng/FoL}}, it struggles when extreme viewpoint changes are compounded by severe occlusions (as illustrated in Fig.~\ref{fig:intro}(a)), due to the lack of explicit fine-grained reliability modeling for occlusion resilience, and incurs inefficiencies from static re-ranking mechanisms. 
We argue that robust global similarity should transcend simple candidate filtering to actively guide dynamic local matching; conversely, local matching reliability should provide implicit guidance to refine global representations. Unlike existing approaches that treat these stages as disconnected components, FoL++ bridges this gap by integrating global priors, local details, and reliability assessments into a unified framework where they mutually reinforce one another.

To address these challenges, FoL++ introduces an end-to-end unified framework. This framework features a Reliability Estimation Branch (REB) to generate spatial reliability maps that guide local matching, employs an Adaptive Candidate Scheduler for dynamic resource allocation during re-ranking, and utilizes a fusion mechanism to combine global and local evidence in the final similarity computation. The REB operates in parallel with the DINOv2 backbone. The Adaptive Candidate Scheduler dynamically adjusts the candidate pool size based on global similarity while applying reliability-aware re-ranking through local feature weights. This cross-stage fusion integrates these elements without requiring separate training phases. Our principal contributions include: the introduction of $\mathcal{L}_{SA}$ and $\mathcal{L}_{SCE}$ to model discriminative local regions, while also generating weakly supervised pseudo-correspondences from global embeddings without pixel-level annotations to improve re-ranking performance; the proposal of the REB, which generates spatial reliability maps to explicitly model occlusion-resistant regions; the development of the Adaptive Candidate Scheduler for dynamic re-ranking, which reduces computational costs by 40\% while maintaining high accuracy; and the establishment of a cross-stage fusion mechanism that integrates global and local evidence into a unified similarity score, addressing the traditional separation of these components.

Our FoL++ introduces the following additional contributions (extending our prior FoL):
\begin{itemize}
    \item \textbf{Robust Discriminative Region Modeling}: We propose the Reliability Estimation Branch to generate spatial reliability maps that explicitly model occlusion resistance. Guided by Spatial Alignment Loss and Saliency Contrast Enhancement Loss, it prioritizes discriminative regions in low-texture urban environments and under extreme viewpoints, enhancing localization accuracy by focusing on robust feature correspondences.

    \item \textbf{Adaptive Candidate Scheduling for Efficient Re-ranking}: We introduce the Adaptive Candidate Scheduler, which dynamically adjusts candidate pool sizes based on global similarity confidence and weights local matches using spatial reliability. This mechanism enables cross-stage fusion of global and local evidence, optimizing the matching system. It significantly reduces computational costs and improves inference speed by 40\% without sacrificing retrieval accuracy (see Fig.~\ref{fig:intro}(b)).

    \item \textbf{Unified Efficient Architecture and SOTA Performance}: We propose a streamlined end-to-end framework that replaces FoL's module-dependent attention layer with the lightweight REB and generates weakly supervised pseudo-correspondences. Finally, it integrates global and local features into a unified similarity score. Extensive experiments across seven benchmarks demonstrate that FoL++ achieves state-of-the-art performance (see Fig.~\ref{fig:intro}(c)), while reducing GPU memory usage by up to 24\% compared to the original FoL (see Fig.~\ref{fig:intro}(b)).
    
\end{itemize}

\section{Related work}
\label{sec:related work}
The following section reviews recent advances in VPR, covering one-stage and two-stage approaches as well as methods that incorporate reliability-aware local feature learning.

\subsection{One-Stage VPR}

One-stage VPR methods generate global image representations by aggregating local features, either handcrafted (e.g., SURF~ \citep{bay2006surf}) or learned from deep networks such as CNNs, MLPs, and transformers~ \citep{sunderhauf2015performance,noh2017large,naseer2017semantics,cao2020unifying}. NetVLAD~ \citep{netvlad}, a CNN-based approach, introduces a trainable VLAD layer for end-to-end feature aggregation, significantly improving retrieval performance.
Benefiting from the rapid development of deep learning, a significant body of work has begun to rely on deep features to solve the VPR problem, while also embedding feature aggregation modules into the model architecture to enhance representation ability~ \citep{jin2017learned,leyva2023data}.
MixVPR \citep{mixvpr} learns a holistic global descriptor by iteratively mixing CNN feature maps with a lightweight, all-MLP architecture, achieving highly competitive performance with significant efficiency.
Transformer-based BoQ~ \citep{boq} employs learnable global queries with cross-attention to dynamically aggregate features, demonstrating strong feature extraction capabilities.
Unlike the contrastive learning methods discussed above, another family of one-stage approaches treats VPR as a classification problem. CosPlace~ \citep{cosplace} and EigenPlaces~ \citep{eigenplaces} assign images to location and view-based classes and train with categorical cross-entropy loss. Divide\&Classify~ \citep{Divide_classify} avoids visual ambiguity by partitioning the map into non-adjacent cells, and can be integrated with retrieval pipelines for better performance.

With the advent of visual foundation models (VFMs), several methods have adopted DINOv2~ \citep{dinov2} as the backbone for VPR, achieving state-of-the-art performance~ \citep{superplace,megaloc}.
AnyLoc~ \citep{anyloc} achieves strong zero-shot universal VPR by aggregating features from the DINOv2 model, though its performance is inherently limited by the lack of fine-tuning for the VPR task.
SALAD~ \citep{salad} reinterprets NetVLAD's~ \citep{netvlad} soft-assignment as an optimal transmission problem, aggregating local features from DINOv2. SuperVLAD~ \citep{supervlad} improves domain adaptability by removing cluster centers and reducing clusters in NetVLAD. SALAD-CM~ \citep{izquierdo2024close} enhances geographic awareness through CliqueMining, improving performance on dense datasets.
ImAge \citep{image} eliminates the need for an explicit aggregator by inserting learnable aggregation tokens into the transformer backbone, leveraging the inherent self-attention mechanism to implicitly generate robust global descriptors. SAGE \citep{sage} dynamically constructs a geo-visual affinity graph during training for adaptive hard sample mining, and introduces a lightweight soft probing module to enhance discriminative local features before the final aggregation.
However, these methods remain vulnerable to perceptual aliasing, as they rely solely on global features while neglecting spatial cues critical for precise localization. In contrast, our FoL++ explicitly models discriminative spatial regions, which generates spatial confidence maps to focus attention on viewpoint-stable, informative areas.

\subsection{Two-Stage VPR}

Recent advancements in VPR have increasingly embraced two-stage retrieval strategies~ \citep{hausler2020hierarchical,berton2021viewpoint,sferrazza2025match,structvpr,structvpr++}, where database images are first ranked based on global feature similarity, followed by a re-ranking step utilizing spatially encoded local features. While methods like Patch-NetVLAD~ \citep{patchvlad} leverage multi-scale patches for re-ranking and TransVPR~ \citep{transvpr} employs transformer-based attention masks to filter discriminative regions, they suffer from inefficiencies in dense feature computation or rigid Top-K selection mechanisms.
AANet~ \citep{lu2023aanet} leverages a hierarchical approach that combines global feature aggregation for candidate retrieval with dynamic local feature alignment (DALF) for efficient re-ranking.
DHE-VPR~ \citep{lu2024deep} uses a transformer-based deep homography network with REI loss for efficient geometric verification.
R2Former~ \citep{r2former} unifies retrieval and re-ranking via feature correlation analysis, yet its fixed pipeline lacks adaptability to dynamic environments. SelaVPR~ \citep{selavpr} integrates adapters into DINOv2 for local feature extraction but introduces additional computational overhead. Revisit Anything~ \citep{garg2024revisit} introduces segment-based retrieval using SuperSegments from SAM~ \citep{kirillov2023segment}, enabling partial image matching under viewpoint variations. EffoVPR~ \citep{effovpr} leverages a foundation model's internal CLS token for global retrieval and its self-attention layers for re-ranking.
SelaVPR++ \citep{selavpr++} efficiently fine-tunes foundation models through lightweight parallel adapters and introduces a novel retrieval paradigm combining binary feature initial filtering with floating-point feature re-ranking. By designing a unified training protocol across multiple datasets, it achieves better overall efficiency and performance.
However, existing approaches suffer from (1) inefficient static candidate scheduling and (2) inadequate spatial modeling under extreme viewpoint changes and occlusions. FoL++ addresses these challenges with a reliability-aware paradigm, introducing a lightweight Reliability Estimation Branch to identify occlusion-resistant regions and an Adaptive Candidate Scheduler to optimize re-ranking depth based on similarity confidence and reliability.

\subsection{Reliability-Aware Local Feature Learning}

Local feature learning in computer vision continues to grapple with challenges posed by occlusions, low-texture regions, and extreme viewpoint variations~ \citep{lu2025feature,bellavia2022sift,liu2024descriptor}. Early work explored the use of reliability maps, computing metrics such as Average Precision (AP) to pinpoint highly discriminative regions~ \citep{r2d2}. Subsequent advancements integrated attention mechanisms for adaptive feature weighting~ \citep{wang2023attention}, while some methods employed semantic segmentation to suppress unreliable features (e.g., sky and moving objects), albeit at the cost of increased deployment complexity~ \citep{xue2023sfd2}. More recent approaches have sought to enhance matching precision and efficiency by modeling spatial structures through multi-relational graphs~ \citep{zhang2024mesa}, capturing high-level context via topic modeling~ \citep{giang2024topicfm+}, or reformulating dense geometric matching with paired masked image modeling~ \citep{zhu2023pmatch}. Other works have prioritized computational efficiency in dense matching but lacked explicit occlusion handling during reliability estimation~ \citep{lindenberger2023lightglue,potje2024xfeat}.
Drawing inspiration from keypoint reliability estimation techniques in SLAM and Structure-from-Motion (SfM)~ \citep{potje2024xfeat}, FoL++ incorporates a lightweight Reliability Estimation Branch that operates in parallel with the DINOv2 backbone.

\section{Proposed Method}

\begin{figure*}[htp]
    \centering
    \includegraphics[width=1\linewidth]{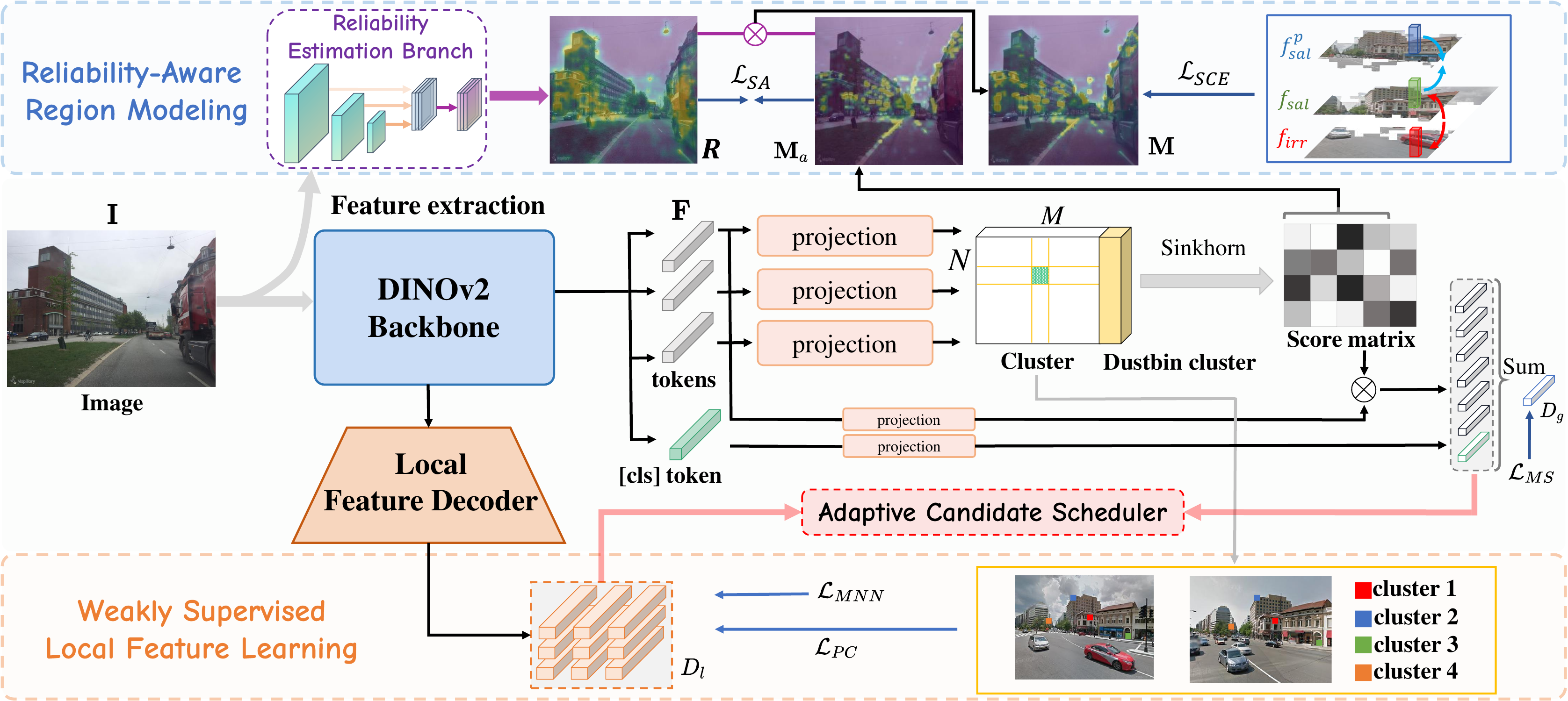}
    \caption{Overview of the FoL++ training pipeline. The DINOv2 backbone extracts patch tokens, which are aggregated via Sinkhorn optimal assignment to form a robust global descriptor $D_g$. In parallel, a lightweight decoder recovers high-resolution local features $D_l$. Meanwhile, the Reliability Estimation Branch generates a spatial reliability map $\mathbf{R}$, which is fused with the global assignment mask $\mathbf{M}_a$ to yield the final discriminative mask $\mathbf{M}$. Local features are optimized via weakly supervised pseudo-correspondences. The entire framework is trained end-to-end by jointly minimizing $\mathcal{L}_{SA}$, $\mathcal{L}_{SCE}$, $\mathcal{L}_{PC}$, $\mathcal{L}_{MS}$.}
    \label{fig:method}
    % \vspace{-0.3cm}
\end{figure*}

In this section, we introduce the FoL++ framework for two-stage VPR. At the core of our reliability-aware paradigm is the robust discriminative region modeling: we propose a lightweight Reliability Estimation Branch (REB) to generate spatial reliability maps that explicitly model occlusion resistance. To optimize this representation, two losses ($\mathcal{L}_{SA}$ and $\mathcal{L}_{SCE}$) are introduced to enhance local contrast and discriminative region learning (see Sections~\ref{sec:arch} and~\ref{sec:rdr}). Furthermore, we employ a weakly supervised pseudo-correspondence module that automatically generates pixel-level supervision by mining spatially consistent feature pairs from aggregation clusters (see Section~\ref{sec:wsl}). To boost inference efficiency and accuracy, our Efficient Re-ranking module leverages the discriminative region mask as a priori cue (see Section~\ref{sec:erd}), while the Adaptive Candidate Scheduler dynamically integrates global and local similarities (see Section~\ref{sec:adaptive_fusion}). The framework is optimized end-to-end by a unified total loss function (see Section~\ref{sec:tlf}).

\subsection{Network Architecture}
\label{sec:arch}

As shown in Figure~\ref{fig:method}, an input image is first processed by a backbone network to extract features, which are then fed into two branches to generate global and local descriptors, respectively. In particular, the global descriptor is obtained via feature aggregation, while the local branch employs a lightweight decoder to produce high-resolution local details for subsequent re-ranking.

\noindent\textbf{Feature Extraction.}
Following recent works~ \citep{keetha2023anyloc}, we adopt the transformer-based vision foundation model DINOv2~ \citep{dinov2} as our feature extractor. Given an input image $\mathbf{I} \in \mathbb{R}^{h \times w \times 3}$, the extractor produces a feature map $\mathbf{F}$ comprising $n = \frac{h\times w}{14^2}$ tokens $\mathbf{t}_i \in \mathbb{R}^{D}$, for $i = 1,\dots,n$. This token set includes a learnable [cls] token that captures global context, as well as patch tokens that encode local details.

\textbf{Global Feature Aggregation.}
To construct a compact global descriptor $\mathbf{D}_g$ from the token set $\mathbf{F}$, we adopt the optimal assignment strategy proposed by SALAD~\citep{salad}, based on the Sinkhorn algorithm~\citep{cuturi2013sinkhorn}. Tokens are first assigned to $M$ salient clusters using a score matrix $\mathbf{S}$, where entries $s_{i,j} = \exp(\mathbf{w}_j^\top \mathbf{t}_i + b_j)$ represent the initial similarity between token $\mathbf{t}_i$ and cluster $j$. By applying the Sinkhorn algorithm to $\mathbf{S}$, we obtain the optimal assignment probability matrix $\mathbf{P}'$, where its element $p'_{i,j}$ denotes the probability of assigning the $i$-th token to the $j$-th cluster. An additional dustbin cluster $(j=M+1)$ is used to exclude irrelevant tokens.

Let $n$ denote the number of patch tokens, with $n = H_p \times W_p$, where $H_p$ and $W_p$ are the patch-grid height and width, respectively. We first compute a token-wise saliency score by summing the assignment probabilities over all $M$ salient clusters:
\begin{equation}
\tilde{\mathbf{M}}_a(i) = \sum_{j=1}^{M} p'_{i,j}, \quad i = 1,\dots,n.
\end{equation}
Note that $\sum_{j=1}^{M} p'_{i,j} = 1 - p'_{i,M+1}$, which reflects the probability that token $i$ belongs to salient clusters rather than the dustbin. We then reshape $\tilde{\mathbf{M}}_a \in \mathbb{R}^{n}$ into the patch-grid dimensions to obtain the final spatial mask $\mathbf{M}_a \in \mathbb{R}^{H_p \times W_p}$. $\mathbf{M}_a$ serves as a attention mechanism to focus on discriminative regions by suppressing irrelevant noise.

\noindent\textbf{Local Feature Decoder.}
The local feature decoder, composed of two deconvolution layers and a ReLU activation, is lightweight and focuses on restoring fine-grained local details.

\begin{figure}[htp]
    \centering
    \includegraphics[width=1\linewidth]{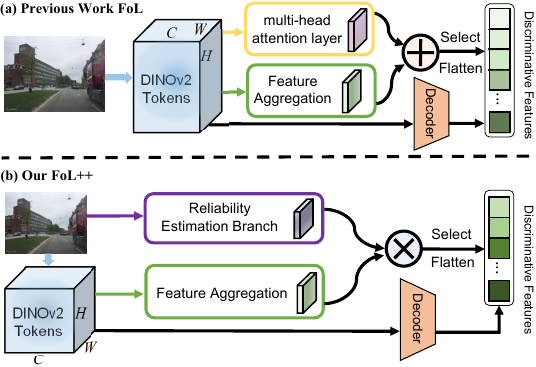}
    \caption{Architectural comparisons between our previous work FoL and our proposed FoL++. In FoL, the core design consists of a multi-head attention layer combined with a feature aggregation branch. In contrast, FoL++ replaces the multi-head attention layer with a lightweight Reliability Estimation Branch.}
    \label{fig:FoL++_REB}
    \vspace{-0.3cm}
\end{figure}

\subsection{Reliable Discriminative Region Modeling}
\label{sec:rdr}
In FoL++, we design two losses, $\mathcal{L}_{SA}$ and $\mathcal{L}_{SCE}$, to model reliable discriminative regions.
We seek to construct discriminative regions by utilizing the spatial information capturing capabilities readily available in feature extraction and global feature aggregation of VPR pipeline.

To improve resource efficiency and matching reliability, we discard $\mathbf{M}_{e}$ from our original FoL~ \citep{FoL}, which is resource-intensive and less reliable. Instead, we introduce a lightweight Reliability Estimation Branch (REB) running parallel to the DINOv2 backbone. Figure~\ref{fig:FoL++_REB} illustrates the architectural distinctions between our previous work FoL and FoL++. Specifically, FoL relies on a multi-head attention layer and a feature aggregation branch, whereas FoL++ employs REB alongside feature aggregation. Inspired by keypoint reliability estimation techniques from SLAM and SfM~ \citep{r2d2,xue2023sfd2}, REB first merges intermediate representations at three scales (1/8, 1/16, and 1/32) by bilinear upsampling and projection to a common resolution of $H/8 \times W/8 \times 64$, followed by element-wise summation.
Following efficient designs~ \citep{detone2018superpoint,potje2024xfeat}, we adopt this resolution to optimize the trade-off between spatial detail and efficiency.
A convolutional fusion block, composed of three basic layers, then produces the final feature representation.
An additional convolutional block regresses a spatial reliability map $\mathbf{R}\in\mathbb{R}^{H/8\times W/8}$, where each element represents the unconditional probability that the corresponding local feature can be matched reliably.
Specifically, the salient region mask $\mathbf{M}_{a}$, derived from the feature aggregation branch as detailed in Sec.~\ref{sec:arch}, is integrated with $\mathbf{R}$ to produce the final discriminative mask. Consistent with the robust nature of salient features against appearance variations~\citep{potje2024xfeat,cadar2024leveraging}, we align $\mathbf{R}$ to the spatial dimensions of $\mathbf{M}_{a}$ via interpolation and fuse them via element-wise multiplication, yielding the final discriminative mask:
\begin{equation}
\mathbf{M} = \mathbf{R} \odot \mathbf{M}_{a}.
\end{equation}

\noindent\textbf{Extraction-Aggregation Spatial Alignment Loss.}
We propose Extraction-Aggregation Spatial Alignment Loss (SAL) to achieve mutually reinforcing self-supervised learning by aligning the regions $\mathbf{R}$ from the feature extractor with the regions $\mathbf{M}_{a}$ preserved by the feature aggregator. Specifically, we align $\mathbf{M}_{a}$ and $\mathbf{R}$ via the Kullback-Leibler divergence~ \citep{van2014renyi}:
\begin{equation}
\mathcal{L}_{SA} = \sum_{i=1}^{h\times w}\mathbf{M}_{a}^{i}\log\frac{\mathbf{M}_{a}^{i}}{\mathbf{R}^{i}} + \sum_{i=1}^{h\times w}\mathbf{R}^{i}\log\frac{\mathbf{R}^{i}}{\mathbf{M}_{a}^{i}},
\end{equation}
where $i$ indexes over the $h\times w$ spatial positions. In practice, we found that direct alignment is challenging due to the significant distributional differences between $\mathbf{M}_{a}$ and $\mathbf{R}$.

Therefore, we smooth $\mathbf{M}_a$ and $\mathbf{R}$. Taking $\mathbf{M}_a$ as an example, each element is clipped at the 90th percentile of its corresponding row:
\begin{equation}
\mathbf{M}_{a}^{(i,j)} = \min\left(\mathbf{M}_{a}^{(i,j)}, \, q_{0.9}\!\left(\mathbf{M}_{a}^{(i,:)}\right)\right),
\end{equation}
where $q_{0.9}\!\left(\mathbf{M}_{a}^{(i,:)}\right)$ denotes the 90th percentile value of the $i$-th row in $\mathbf{M}_a$. In other words, values in the top 10\% of each row are truncated to the row-wise 90th-percentile threshold. 
Conceptually analogous to regularization strategies that prevent over-confidence (such as label smoothing \citep{wall2026winsor,zhou2026maxsup}), our experiments reveal that flattening the top-tier values encourages the network to distribute its attention more evenly across a broader set of reliable structural features rather than over-fitting to isolated points. Ultimately, this robust percentile-based truncation helps mitigate the influence of extreme activations and stabilizes the subsequent Kullback-Leibler divergence computation.

\noindent\textbf{Saliency Contrast Enhancement Loss.}
We further introduce a Saliency Contrast Enhancement Loss (SCEL) to encourage the network to generate regions that are both reliable (i.e., produce consistent and robust attention across images) and discriminative (i.e., are clearly differentiated from irrelevant interference). In our context, \textit{salient regions} refer to the high-utility parts of the image that contain key structures and meaningful information (such as buildings or landmarks critical for visual place recognition), while \textit{irrelevant regions} comprise the less informative or distracting areas (such as sky, vegetation, or transient objects).

\begin{figure}[!htbp]
    \centering
    \includegraphics[width=0.95\linewidth]{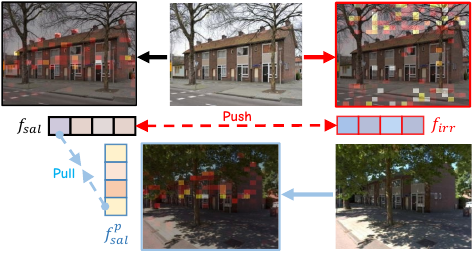}
    \caption{
        Illustration of the $\mathcal{L}_{SCE}$. The salient region descriptor $f_{sal}$ is pulled closer to the positive sample's salient descriptor $f^{p}_{sal}$ while being pushed away from the irrelevant region descriptor $f_{irr}$.
    }
    \label{fig:CEL}
    % \vspace{-0.3cm}
\end{figure}

To achieve this, we first reshape the feature map $\mathbf{F}$, which contains the patch tokens extracted by the backbone, into a tensor $\mathcal{F} \in \mathbb{R}^{h\times w\times d}$ matching the spatial dimensions of the mask $\mathbf{M}$. 

Conceptually, to contrast different regions, we extract a condensed 1D representation for each region type, which we term a \textit{region descriptor}. A region descriptor summarizes the holistic visual characteristics of either the salient or irrelevant areas within a single image. Specifically, the salient region descriptor $f_{sal}$ is obtained by performing an L2-normalized, mask-weighted spatial pooling over $\mathcal{F}$:
\begin{equation}
f_{sal} = \frac{\sum^{h\times w}_{i}\mathbf{M}(i)\cdot \mathcal{F}(i)}{\left\|\sum^{h\times w}_{i}\mathbf{M}(i)\cdot \mathcal{F}(i)\right\|},
\label{eq:2}
\end{equation}
where $\|\cdot\|$ denotes the L2 norm. Similarly, the irrelevant region descriptor $f_{irr}$, which summarizes the background distraction features, is obtained by inverting the mask:
\begin{equation}
f_{irr} = \frac{\sum^{h\times w}_{i}(1-\mathbf{M}(i))\cdot \mathcal{F}(i)}{\left\|\sum^{h\times w}_{i}(1-\mathbf{M}(i))\cdot \mathcal{F}(i)\right\|}.
\label{eq:3}
\end{equation}

Next, leveraging the triplet loss framework~\citep{triplet}, the $\mathcal{L}_{SCE}$ is defined as:
\begin{equation}
 \mathcal{L}_{SCE} = \max\Bigl(0,\, 1-\bigl(f_{sal}\cdot f^{p}_{sal} - f_{sal}\cdot f_{irr}\bigr)\Bigr),
\end{equation}
where $f^{p}_{sal}$ is the salient region descriptor of a positive sample (i.e., another image from the same scene), as shown in Figure~\ref{fig:method}. Note that to ensure the network only optimizes the region mask $\mathbf{M}$ without altering the extracted features $\mathcal{F}$, the gradient of $\mathcal{F}$ is truncated when computing $\mathcal{L}_{SCE}$.

As illustrated in Figure~\ref{fig:CEL}, the $\mathcal{L}_{SCE}$ encourages the model to “pull” $f_{sal}$ closer to $f^{p}_{sal}$ while “pushing” $f_{sal}$ away from $f_{irr}$, thereby reinforcing discriminative feature activation.
Driven by $\mathcal{L}_{SCE}$, the network is encouraged to generate a mask $\mathbf{M}$ that consistently highlights reliable and discriminative salient regions while clearly suppressing irrelevant interference. As a result, this enhances both global feature aggregation and local matching accuracy.

\subsection{Weakly Supervised Local Feature Learning}
\label{sec:wsl}

Due to the lack of pixel-level correspondence labels in VPR datasets, existing two-stage methods can usually only supervise the final matching results, such as R2Former~\citep{r2former}'s classification loss and CricaVPR~\citep{cricavpr}'s mutual nearest neighbor loss. In FoL++, we exploit the clustering results of local features in the global aggregation stage to construct pseudo-correspondence ground truth, thereby enabling pixel-level supervision for local feature matching.
\begin{algorithm}[htbp]
\caption{Pseudo-correspondence Ground Truth Construction}
\label{alg:algorithm}
\begin{algorithmic}[1]
\State \textbf{Input:} Anchor image $I_p$ features $\mathbf{F}_p$, region map $\mathbf{M}$, assignment matrix $\mathbf{S}_p$; Positive image $I_{p'}$ features $\mathbf{F}_{p'}$, assignment matrix $\mathbf{S}_{p'}$
\State \textbf{Parameters:} $thr1 = 0.8$, $thr2 = 0.5$, $N = 12$
\State \textbf{Initialize:} $\mathcal{P} = \emptyset$. Rank patches in $\mathbf{F}_p$ descendingly via $\mathbf{M}$.
\While{$|\mathcal{P}| < N$ \textbf{and} patches remain}
    \State Pop top patch feature $f_p$ (index $p$) and get its cluster $C_{f_p}$ via $\mathbf{S}_p$
    \State Find $\mathbf{F}_{p'}$ candidates matching $C_{f_p}$ via $\mathbf{S}_{p'}$
    \State Let $\{f_{p'_{1th}}, f_{p'_{2th}}\}$ be the top-2 candidates most similar to $f_p$
    \State Let similarity ratio $r = \frac{sim(f_p, f_{p'_{2th}})}{sim(f_p, f_{p'_{1th}})}$
    \If{$sim(f_p, f_{p'_{1th}}) > thr1$ \textbf{and} $r < thr2$}
    \State $\mathcal{P} \leftarrow \mathcal{P} \cup \{(p, p'_{1th})\}$
\EndIf
\EndWhile
\State \textbf{Return:} $\mathcal{P}$
\end{algorithmic}
\end{algorithm}

\begin{figure}[!htbp]
\setlength{\abovecaptionskip}{-0.02cm}
    \centering
    \includegraphics[width=1\linewidth]{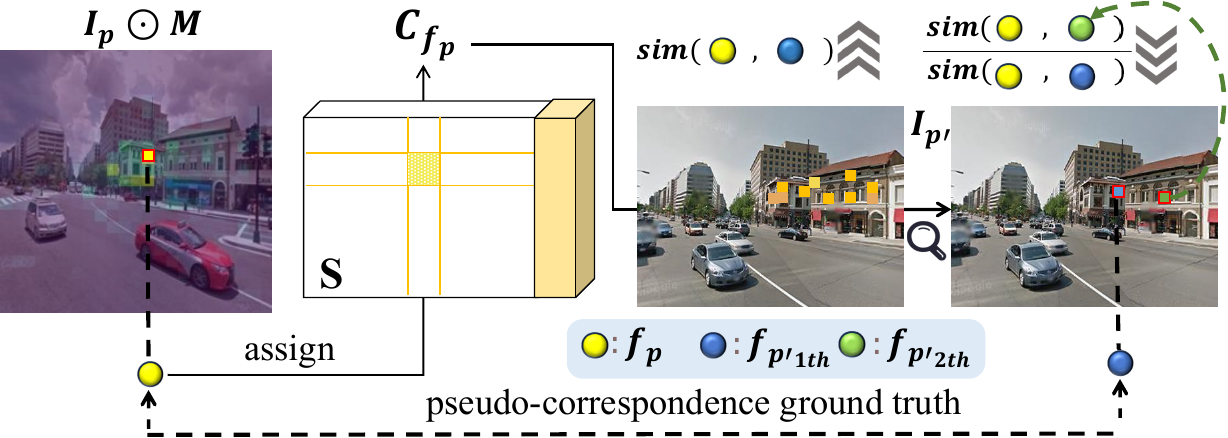}
    \caption{
        Flowchart of the Weakly Supervised Local Feature Learning.
    }
    \label{fig:ws-net}
    % \vspace{-0.3cm}
\end{figure}

\noindent\textbf{Pseudo-correspondence Ground Truth Construction.}
As shown in Figure~\ref{fig:ws-net}, each patch is assigned to a cluster during global feature aggregation, which implicitly encodes patch correspondences. Based on this observation, we build pseudo correspondences by selecting reliable patches from the anchor image and matching them with cluster-consistent candidates from the positive image. The detailed procedure is summarized in Algorithm~\ref{alg:algorithm}.

Specifically, given an anchor image $I_p$, its discriminative region map $\mathbf{M}$, and its assignment matrix $\mathbf{S}_p$, we first rank patches by their activation values in $\mathbf{M}$. Then, we select the patch feature $f_p$ with the highest activation value and obtain its cluster index $C_{f_p}$ from $\mathbf{S}_p$. Next, using the positive image's assignment matrix $\mathbf{S}_{p'}$, we search for all patches assigned to $C_{f_p}$ in the positive image $I_{p'}$ and select the top-2 candidates, denoted by $f_{p'_{1th}}$ and $f_{p'_{2th}}$, according to their similarity to $f_p$. If the best candidate satisfies the absolute similarity threshold and the ratio test, we accept the spatial index pair $(p, p'_{1th})$ as a valid pseudo-correspondence. This filtering strategy retains patches that are both distinctive and reliable, reducing ambiguous matches and improving intra-class feature consistency.

To ensure the reliability of the generated pseudo-labels, we employ a dual-filtering mechanism within the selection process. Specifically, the absolute similarity threshold ($thr1$) acts as a relevance gate, enforcing high visual coherence to discard obviously unrelated or low-quality matches. Complementing this, the ratio test (controlled by $thr2$) functions as a uniqueness check, conceptually inspired by Lowe’s classic strategy in SIFT feature matching~\citep{SIFT}. By requiring the similarity ratio between the best and second-best matches to fall below $thr2$, this criterion effectively eliminates ambiguous correspondences arising from repetitive architectural patterns. This combination guarantees that the retained matches are both visually consistent and discriminatively unique, providing robust supervision signals despite the absence of manual annotations.

\noindent\textbf{Weakly Supervised Pseudo-correspondence Loss.}

We obtain $(D_{l}^{p}, D_{l}^{p^{'}})$ by sampling the network output dense local features $D_{l}$ using the pseudo-correspondence $(p, p^{'})$. From this, $\mathcal{L}_{PC}$ can be defined as:
\begin{equation}
\mathcal{L}_{PC}=\frac{\sum exp(sim(f_{p_{i}},f_{p^{'}_{i}}))\cdot (1-sim(D_{l}^{p_{i}},D_{l}^{p^{'}_{i}})))}{\sum exp(sim(f_{p_{i}},f_{p^{'}_{i}}))}.
\end{equation}
Since no true ground truth is used, we use similarity as a confidence for evaluating the quality of labels to mitigate the negative impact of inaccurate labels on model training, hence $\mathcal{L}_{PC}$ is a weakly supervised loss.

\subsection{Efficient Re-ranking with Discriminative Region Guidance}
\label{sec:erd}

We propose to accelerate and enhance the re-ranking process using the local regions mask $\mathbf{M}$ modeled in previous section as a priori information for local matching.

First, we convert $\mathbf{M}$ into a binary matrix, where the top \(k\) positions in $\mathbf{M}$ are set to 1, and others to 0. This can be expressed as:

\begin{equation}
\mathbf{M}_{\text{bin}} =
\begin{cases}
1 & \text{if index} \in \text{top } k \text{ of } \mathbf{M} \\
0 & \text{otherwise},
\end{cases}
\end{equation}
where \(k\) is set to the top $40\%$.

Then, we interpolate \(\mathbf{M}_{\text{bin}}\) and select the parts of the dense local features $D_{l}$ where the mask is 1:

\begin{equation}
D_{l}^{\mathbf{M}} = D_{l}[\mathbf{M}_{\text{bin}} = 1]
\end{equation}

Finally, we only perform nearest-neighbor local feature matching on $D_{l}^{\mathbf{M}}$, which reduces the probability of mismatching and improves the efficiency of matching due to the narrowed matching range.

\subsection{Adaptive Candidate Scheduler and Cross-Stage Fusion}
\label{sec:adaptive_fusion}

The Adaptive Candidate Scheduler framework integrates global and local visual information through Dynamic Top-k Candidate Selection, Reliability-Aware Local Matching (RALM), and Similarity Computation (SC). DCS dynamically adjusts the candidate pool size based on global similarity, reducing computations while preserving retrieval accuracy in complex scenarios. RALM refines local matches using spatial reliability maps, weighting mutual nearest neighbors based on the reliability of corresponding regions in both query and candidate images. SC then fuses global and local scores through weighted summation, enabling a unified decision process that surpasses traditional methods relying on independent matching stages \citep{patchvlad, selavpr}.

\begin{figure}[!htbp]
\setlength{\abovecaptionskip}{-0.02cm}
    \centering
    \includegraphics[width=1\linewidth]{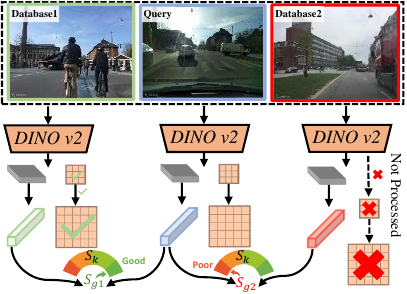}
\caption{
    Illustration of Dynamic Top-k Candidate Selection. The query image (center) is compared with two database images using global feature similarity. The left image, with a similarity \(S_{g1}\) above the threshold \(S_k\), is deemed a "Good" match and undergoes local processing, while the right image, with a similarity \(S_{g2}\) below \(S_k\), bypasses local processing.
}
    \label{fig:DCS}
    \vspace{-0.5cm}
\end{figure}

\subsubsection{Dynamic Top-k Candidate Selection (DCS)}
\label{subsec:acs}

Conventional methods often employ a fixed-size candidate pool for re-ranking~ \citep{patchvlad,transvpr,selavpr,effovpr}. In contrast, DCS dynamically adjusts the number of candidates based on the confidence of the global similarity scores. As illustrated in Figure~\ref{fig:DCS}, the query image (center) is compared with two database images using global feature similarity. The left image, with a similarity \(S_{g1}\) above the threshold \(S_k\), is deemed a "Good" match and undergoes local processing, while the right image, with a similarity \(S_{g2}\) below \(S_k\), bypasses local processing.
Specifically, the candidate pool size is determined as follows:
\begin{equation}
k = \min\left\{\left\lfloor \alpha \cdot P(S_q) \cdot (k_{\max} - k_{\min}) + k_{\min} \right\rfloor,\, k_{\max} \right\},
\label{eq:acs}
\end{equation}
where \( S_g = \{ S_{g1}, S_{g2}, \dots, S_{g k_{\max}} \} \) represents the set of global similarity scores between the query and gallery images. The adaptive threshold \( S_q \) is computed as the mean of the top-\( k' \) highest similarity scores:
\begin{equation}
S_q = \frac{1}{k'} \sum_{i=1}^{k'} S_{g(i)},
\end{equation}
where \( S_{g(i)} \) denotes the \( i \)-th highest similarity score in \( S_g \), and \( k' \) is set to 60 as a fixed value. The percentile function \( P(S_q) \) is defined as:
\begin{equation}
P(S_q) = \frac{1}{k_{\max}} \sum_{i=1}^{k_{\max}} \mathbb{I}(S_{g(i)} \leq S_q),
\end{equation}
where \( \mathbb{I}(\cdot) \) is an indicator function that returns 1 if \( S_{g(i)} \leq S_q \) and 0 otherwise. The scaling factor \( \alpha \) controls the sensitivity of candidate pool size adjustment, while \( k_{\min} \) and \( k_{\max} \) define the lower and upper bounds of the candidate pool (30 and 100, respectively).
By leveraging percentile-based ranking and robust score selection, DCS dynamically optimizes the candidate pool size based on the confidence of global similarity scores, ensuring a trade-off between computational efficiency and retrieval accuracy.

\subsubsection{Reliability-Aware Local Matching (RALM)}
\label{subsec:local_matching}
RALM improves local similarity computation by weighting matches with spatial reliability. Local similarity is measured by counting the number of mutual nearest neighbor matches \citep{selavpr}. Given a query image \(q\) and a candidate image \(c\), let \(\boldsymbol{f}^l_q(i)\) be the \(i\)-th local feature in the query image and \(\boldsymbol{f}^l_c(j)\) be the \(j\)-th local feature in the candidate image. The local feature similarity \(\boldsymbol{s}_{qc}(i,j)\) between these features is computed, and the mutual nearest neighbor match set \(\mathcal{M}\) is defined as
\begin{equation}
\begin{split}
\mathcal{M} = \Bigl\{ (u,v) \;\Big|\; 
& u = \arg\max_{i}\boldsymbol{s}_{qc}(i,v), \\
& v = \arg\max_{j}\boldsymbol{s}_{qc}(u,j) \Bigr\}.
\end{split}
\label{eq:mn}
\end{equation}
To further refine the local similarity measure, a spatial reliability map \(\mathbf{R} \in \mathbb{R}^{H/8 \times W/8}\) is computed for each image, where each element represents the unconditional probability that the corresponding local feature is reliably matched. 
The reliability map does not replace local features in matching; instead, it modulates the contribution of each matched pair by emphasizing stable regions and suppressing unreliable ones.
Since the resolution of \(\mathbf{R}\) is lower than that of the local feature map, it is first upsampled via interpolation. The improved local similarity score is then defined as
\begin{equation}
S_l = \sum_{(u,v) \in \mathcal{M}} \sqrt{R_q(u)\, R_c(v)},
\label{eq:improved_Sl}
\end{equation}
where \(R_q(u)\) and \(R_c(v)\) denote the upsampled reliability values at the locations of the \(u\)-th and \(v\)-th local features in the query and candidate images, respectively. This weighting mechanism modulates the contribution of established correspondences, penalizing matches in unstable regions while highlighting reliable landmarks.

\subsubsection{Similarity Computation (SC)}
\label{subsec:score_fusion}
The overall image similarity is computed by a lightweight late-fusion strategy that combines global retrieval evidence and reliability-weighted local verification:
\begin{equation}
S_{\text{final}} = \gamma\, S_g + S_l,
\label{eq:final_fusion}
\end{equation}
where $\gamma$ is a weighting factor used to balance the different score scales of the global similarity $S_g$ and the improved local similarity $S_l$. This linear fusion is motivated by the complementary roles of the two scores: $S_g$ provides coarse retrieval cues for candidate ranking, while $S_l$ captures fine-grained geometric consistency during re-ranking. In our implementation, $\gamma$ is selected on the Pitts30k-val.

\subsection{Total Loss Function}
\label{sec:tlf}

Following previous works~ \citep{cricavpr,boq,salad}, we also use the multi-similarity loss~ \citep{multiloss} $\mathcal{L}_{MS}$ with an online hard mining strategy to optimize the global features.
To optimize the local features matching, we also adopt the mutual nearest neighbor local feature loss \(\mathcal{L}_{MNN}\) proposed in SelaVPR \citep{selavpr}.
Combined with the techniques presented in the previous sections, the total loss can be defined as:
\begin{equation}
\mathcal{L} = \mathcal{L}_{MS}+ \mathcal{L}_{MNN}+ \mathcal{L}_{SCE}+ \alpha \mathcal{L}_{SA}+ \beta \mathcal{L}_{PC},
\end{equation}
% where $\alpha$ and $\beta$ are weights that regulate the optimization and are empirically set to $5e-5$ and $1e-4$.
where $\{\alpha, \beta\}$ are hyperparameters to adjust the magnitude of the different losses, consistent with FoL~ \citep{FoL}.

\section{Experiments}
\label{sec:experiment}

\subsection{Benchmarks and Performance Evaluation}

We rigorously assess our models across a diverse array of VPR benchmarks, encompassing urban landscapes, seasonal dynamics, and expansive environments, as detailed in Table~\ref{tab:eval_datasets}. Our primary evaluations focus on Pitts30k-test~ \citep{torii2013visual}, MSLS-val~ \citep{warburg2020mapillary}, Nordland~ \citep{sunderhauf2013we}, Tokyo24/7~ \citep{densevlad}, Nordland$\star$~ \citep{sunderhauf2013we}, Pitts250k~ \citep{torii2013visual}, and Eynsham~ \citep{cummins2009highly,benchmark}.

\textbf{Pitts30k-test}: Derived from Google Street View, this urban dataset includes significant viewpoint changes, moderate environmental variations, and minimal dynamic elements, serving as a standard test set for small-scale scenarios.

\textbf{MSLS-val}: A subset of the extensive Mapillary Street-Level Sequences dataset, containing 18,871 database images and 740 queries. It tests models under diverse conditions including weather, lighting, and dynamic objects.

\textbf{Nordland}: Captured from a train route across four seasons, this dataset focuses on suburban/natural environments with strong seasonal and lighting changes but zero viewpoint variations, specifically assessing robustness to extreme appearance shifts.

\textbf{Tokyo24/7}: Features 75,984 database images and 315 queries. The queries cover three distinct daily times (Day, Sunset, Night), challenging models with extreme illumination, temporal, and moderate viewpoint shifts.

\textbf{Nordland$\star$}: A challenging cross-seasonal evaluation protocol where a subset of 2,760 summer images is used as queries against the entire winter sequence (27,592 reference images), specifically testing the models' capability to handle extreme seasonal appearance shifts.

\textbf{Pitts250k}: The large-scale version of the Pittsburgh dataset (83,952 database images and 8,280 queries), enabling comprehensive evaluation of place recognition performance across expansive urban environments.

\textbf{Eynsham}: A grayscale dataset (23,935 queries and database images) captured along an Oxford route using a car-mounted camera. Its lack of color information and temporal variations provides a stringent benchmark for pure structural and geometric place recognition.

\begin{table}[!t]
\centering
\caption{A summary of benchmark datasets for evaluation.}
\label{tab:eval_datasets}
\setlength{\tabcolsep}{1pt} 
\fontsize{8pt}{9.5pt}\selectfont
\renewcommand{\arraystretch}{1.15}
\begin{tabular}{llrr}
\toprule
\multirow{2}{*}{\textbf{Dataset}} & \multirow{2}{*}{\textbf{Description}} & \multicolumn{2}{c}{\textbf{Number}} \\
\cmidrule(lr){3-4}
 & & \textbf{Database} & \textbf{Queries} \\
\midrule
Pitts30k-test & GSV, urban (subset) & 10,000 & 6,816 \\
MSLS-val & Mapillary subset & 18,871 & 740 \\
Nordland & Train route, seasonal & 27,592 & 27,592 \\
Tokyo24/7 & Urban, day-night & 75,984 & 315 \\
Nordland$\star$ & Train route, seasonal & 27,592 & 2,760 \\
Pitts250k & GSV, urban (large-scale) & 83,952 & 8,280 \\
Eynsham & Oxford, grayscale & 23,935 & 23,935 \\
\bottomrule
\end{tabular}
\vspace{-0.5cm}
\end{table}

We follow standard evaluation protocols\footnote{We organize the evaluation following the Deep Visual Geo-Localization Benchmark~ \citep{berton2022deep}: \url{https://github.com/gmberton/deep-visual-geo-localization-benchmark}.} used in previous works~ \citep{berton2022deep,boq,patchvlad}, assessing performance using Recall@N.
For Pitts30k, Eynsham, Pitts250k, Tokyo24/7, and MSLS, retrieval correctness is determined within a 25-meter threshold. For Nordland, evaluation is performed with a tolerance of $\pm10$ frames, effectively measuring retrieval accuracy under various challenging conditions.

\begin{table*}[t]
    \centering
    \setlength{\tabcolsep}{0.20mm}
    \scriptsize
    \caption{Comparison with SoTA methods on VPR benchmarks. 'Res' indicates that all methods are evaluated at $322 \times 322$ resolution. \textbf{\textcolor{red}{Red bold}} and \textbf{\textcolor{blue}{blue bold}} denote the best and second-best results, respectively. $\dag$ marks two-stage methods.}
    \vspace{-0.1cm}

    \renewcommand{\arraystretch}{1.1}
    \begin{tabular}{@{}l|c||ccc||ccc||ccc||ccc||ccc}
    \toprule
    \multirow{2}{*}{Method} & \multirow{2}{*}{Res}
      & \multicolumn{3}{c||}{Pitts30k-test}
      & \multicolumn{3}{c||}{MSLS-val}
      & \multicolumn{3}{c||}{Nordland}
      & \multicolumn{3}{c||}{Tokyo24/7}
      & \multicolumn{3}{c}{Nordland$\star$} \\ 
    \hhline{~~---------------} 
      & & \small{R@1} & \small{R@5} & \small{R@10}
      & \small{R@1} & \small{R@5} & \small{R@10}
      & \small{R@1} & \small{R@5} & \small{R@10}
      & \small{R@1} & \small{R@5} & \small{R@10}
      & \small{R@1} & \small{R@5} & \small{R@10} \\
    \hline
    NetVLAD$_{\textcolor{blue}{\text{\tiny CVPR'16}}}$ & 322
      & 81.9 & 91.2 & 93.7 & 53.1 & 66.5 & 71.1 & 6.4  & 10.1 & 12.5 & 60.6 & 68.9 & 74.6 & 32.6 & 47.1 & 53.3 \\
    Patch-NetVLAD$_{\textcolor{blue}{\text{\tiny CVPR'21}}}$$^\dag$ & 322
      & 88.7 & 94.5 & 95.9 & 79.5 & 86.2 & 87.7 & 44.9 & 50.2 & 52.2 & 86.0 & 88.6 & 90.5 & 30.8 & 34.5 & 35.4 \\
    CosPlace$_{\textcolor{blue}{\text{\tiny CVPR'22}}}$ & 322
      & 88.4 & 94.5 & 95.7 & 82.8 & 89.7 & 92.0 & 58.5 & 73.7 & 79.4 & 81.9 & 90.2 & 92.7 & 34.4 & 49.9 & 56.5 \\
    TransVPR$_{\textcolor{blue}{\text{\tiny CVPR'22}}}$$^\dag$ & 322
      & 89.0 & 94.9 & 96.2 & 86.8 & 91.2 & 92.4 & 63.5 & 68.5 & 70.2 & 79.0 & 82.2 & 85.1 & 46.9 & 51.6 & 52.5 \\
    SelaVPR$_{\textcolor{blue}{\text{\tiny ICLR'24}}}$ & 322
      & 90.2 & 96.1 & 97.1 & 87.7 & 95.8 & 96.6 & 72.3 & 89.4 & 94.4 & 81.9 & 94.9 & 96.5 & 43.8 & 63.1 & 71.1 \\
    MixVPR$_{\textcolor{blue}{\text{\tiny WACV'23}}}$ & 322
      & 91.5 & 95.5 & 96.3 & 88.0 & 92.7 & 94.6 & 76.2 & 86.9 & 90.3 & 85.1 & 91.7 & 94.3 & 58.4 & 74.6 & 80.0 \\
    EigenPlaces$_{\textcolor{blue}{\text{\tiny ICCV'23}}}$ & 322
      & 92.5 & 96.8 & 97.6 & 89.1 & 93.8 & 95.0 & 71.2 & 83.8 & 88.1 & 93.0 & 96.2 & 97.5 & 54.4 & 68.8 & 74.1 \\
    SelaVPR$_{\textcolor{blue}{\text{\tiny ICLR'24}}}$$^\dag$ & 322
      & 92.8 & 96.8 & 97.7 & 90.8 & 96.4 & 97.2 & 87.3 & 93.8 & 95.6 & 94.0 & 96.8 & 97.5 & 63.0 & 77.6 & 81.4 \\
    SALAD$_{\textcolor{blue}{\text{\tiny CVPR'24}}}$ & 322
      & 92.5 & 96.4 & 97.5 & 92.2 & 96.4 & 97.0 & 89.7 & 95.5 & 97.0 & 94.6 & 97.5 & 97.8 & 76.0 & 89.2 & 92.0 \\
    SALAD-CM$_{\textcolor{blue}{\text{\tiny ECCV'24}}}$ & 322
      & 92.6 & 96.8 & 97.8 & \textbf{\textcolor{red}{94.2}} & 97.2 & 97.4 & \textbf{\textcolor{blue}{96.0}} & \textbf{\textcolor{blue}{98.5}} & \textbf{\textcolor{red}{99.2}} & 96.8 & 97.5 & 97.8 & \textbf{\textcolor{red}{90.7}} & \textbf{\textcolor{blue}{96.6}} & \textbf{\textcolor{blue}{97.5}} \\
    BoQ$_{\textcolor{blue}{\text{\tiny CVPR'24}}}$ & 322
      & 93.7 & 97.1 & 97.9 & 93.8 & 96.8 & 97.0 & 90.6 & 96.0 & 97.5 & \textbf{\textcolor{red}{98.1}} & 98.1 & 98.7 & 81.3 & 92.5 & 94.8 \\
      SuperVLAD$_{\textcolor{blue}{\text{\tiny NIPS'24}}}$ & 322
        & \textbf{\textcolor{red}{95.0}} & \textbf{\textcolor{red}{97.4}} & \textbf{\textcolor{red}{98.2}}
        & 92.2 & 96.6 & 97.4
        & 91.0 & 96.4 & 97.7
        & 95.6 & 97.8 & 98.1
        & 75.1 & 89.5 & 93.0 \\
    EDTFormer$_{\textcolor{blue}{\text{\tiny TCSVT'25}}}$ & 322
      & 93.4 & 97.0 & 97.9 & 92.0 & 96.6 & 97.2 & 88.3 & 95.3 & 97.0 & \textbf{\textcolor{blue}{97.1}} & 98.1 & 98.4 & 73.1 & 86.7 & 90.1 \\
    MegaLoc$_{\textcolor{blue}{\text{\tiny CVPRW'25}}}$ & 322
      & 94.1 & \textbf{\textcolor{blue}{97.3}} & \textbf{\textcolor{red}{98.2}} & 93.5 & 97.0 & 97.8 & 94.2 & 97.9 & 98.6 & 96.5 & 98.1 & \textbf{\textcolor{red}{99.7}} & 88.1 & 95.6 & 97.0 \\

    ImAge$_{\textcolor{blue}{\text{\tiny NIPS'25}}}$ & 322
        & 94.0 & 97.2 & 98.0
        & 93.0 & 97.0 & 97.2
        & 93.2 & 97.6 & 98.6
        & 96.2 & 98.1 & 98.4
        & 84.6 & 95.1 & 97.0 \\

    SAGE $_{\textcolor{blue}{\text{\tiny ICLR'26}}}$ & 322
        & 93.4 & 97.0 & 97.9
        & 93.4 & 97.3 & 97.6
        & 94.1 & 98.0 & \textbf{\textcolor{blue}{98.8}}
        & \textbf{\textcolor{blue}{97.1}} & \textbf{\textcolor{blue}{98.4}} & \textbf{\textcolor{blue}{99.1}}
        & 86.7 & 95.9 & 97.0 \\
        
    \hline
    FoL (global) & 322
      & 93.6 & 96.9 & 97.9 & 92.8 & 96.9 & 97.2 & 83.8 & 92.6 & 95.1 & 96.5 & 98.1 & 98.4 & 74.1 & 88.8 & 92.2 \\
    FoL$^\dag$ & 322
      & 93.9 & 96.9 & \textbf{\textcolor{blue}{98.1}} & 90.1 & 95.7 & 96.9 & 87.9 & 94.8 & 96.6 & \textbf{\textcolor{blue}{97.1}} & 97.8 & 98.7 & 80.8 & 92.0 & 94.7 \\
    \hline
    \rowcolor{lightshade}
    FoL++ (global) & 322
      & 93.8 & 97.1 & \textbf{\textcolor{blue}{98.1}} & 93.2 & \textbf{\textcolor{red}{98.0}} & \textbf{\textcolor{red}{98.2}} & 91.8 & 96.9 & 98.1 & 96.8 & \textbf{\textcolor{blue}{98.4}} & 98.4 & 84.5 & 94.4 & 96.3 \\
    \rowcolor{lightshade}
    FoL++$^\dag$ & 322
      & \textbf{\textcolor{blue}{94.7}} & \textbf{\textcolor{red}{97.4}} & \textbf{\textcolor{blue}{98.1}} & \textbf{\textcolor{blue}{94.1}} & \textbf{\textcolor{blue}{97.8}} & \textbf{\textcolor{blue}{98.1}} & \textbf{\textcolor{red}{96.1}} & \textbf{\textcolor{red}{98.6}} & \textbf{\textcolor{red}{99.2}} & \textbf{\textcolor{red}{98.1}} & \textbf{\textcolor{red}{99.1}} & \textbf{\textcolor{blue}{99.1}} & \textbf{\textcolor{blue}{89.7}} & \textbf{\textcolor{red}{96.7}} & \textbf{\textcolor{red}{97.8}} \\
    \bottomrule
    \end{tabular}
    \label{tab:compare_SOTA}
    % \vspace{-0.3cm}
\end{table*}

\begin{table}[ht]
\vspace{-0.3cm}
\caption{A further performance comparison on more datasets. The Image Resolution Is $224\times224$ in Training and $322\times322$ in Inference. $\dag$ marks two-stage methods.}
\centering
\setlength{\tabcolsep}{1.2pt}
\fontsize{7.8pt}{9.0pt}\selectfont
\renewcommand{\arraystretch}{1.0}
\begin{tabular}{lcccccc}
\toprule
\multirow{2}{*}{Method} & \multicolumn{3}{c}{Pitts250k} & \multicolumn{3}{c}{Eynsham} \\ \cmidrule(lr){2-4} \cmidrule(lr){5-7}
                        & R@1 & R@5 & R@10 & R@1 & R@5 & R@10 \\
\midrule
\multicolumn{7}{c}{\textit{ViT-B Architecture}} \\ 
\midrule
\rowcolor{lightblue}
\multicolumn{7}{l}{\textit{Single-Stage Methods}} \\
CricaVPR$_{\textcolor{blue}{\text{\tiny CVPR'24}}}$  & 95.6 & 98.9 & \textbf{\textcolor{red}{99.5}} & 91.6 & 95.0 & 95.8 \\
SALAD$_{\textcolor{blue}{\text{\tiny CVPR'24}}}$     & 95.1 & 98.5 & 99.1 & 91.6 & 95.1 & 95.9 \\
SALAD-CM$_{\textcolor{blue}{\text{\tiny ECCV'24}}}$ & 95.2 & 98.8 & 99.3 & 91.9 & 95.3 & 96.1 \\
BoQ$_{\textcolor{blue}{\text{\tiny CVPR'24}}}$       & \textbf{\textcolor{red}{96.6}} & \textbf{\textcolor{red}{99.1}} & \textbf{\textcolor{red}{99.5}} & 92.2 & 95.6 & 96.4 \\
EDTFormer$_{\textcolor{blue}{\text{\tiny TCSVT'25}}}$ & 95.9 & 98.8 & 99.3 & 92.2 & 95.5 & 96.4 \\
ImAge$_{\textcolor{blue}{\text{\tiny NIPS'25}}}$ & 96.5 & \textbf{\textcolor{red}{99.1}} & \textbf{\textcolor{red}{99.5}} & 92.0 & 95.4 & 96.1 \\
SAGE $_{\textcolor{blue}{\text{\tiny ICLR'26}}}$ & 95.7 & 98.6 & 99.2 & \textbf{\textcolor{red}{92.3}} & \textbf{\textcolor{red}{95.8}} & 96.5 \\
\rowcolor{lightblue}
\multicolumn{7}{l}{\textit{Two-Stage Methods}} \\
FoL$_{\textcolor{blue}{\text{\tiny AAAI'25}}}$$^\dag$ & 96.1 & \textbf{\textcolor{red}{99.1}} & \textbf{\textcolor{red}{99.5}} & 91.3 & 95.1 & 96.1 \\
SelaVPR++$_{\textcolor{blue}{\text{\tiny TPAMI'25}}}$$^\dag$ & 95.9 & 98.6 & 99.2 & \textbf{\textcolor{red}{92.3}} & 95.6 & 96.5 \\
\rowcolor{lightshade}
FoL++ (global) & 94.6 & 98.3 & 99.0 & 91.4 & 95.4 & 96.3 \\
\rowcolor{lightshade}
FoL++$^\dag$   & 95.7 & 98.6 & 99.2 & 92.1 & 95.7 & \textbf{\textcolor{red}{96.6}} \\
\midrule
\multicolumn{7}{c}{\textit{ViT-L Architecture}} \\ 
\midrule
FoL$_{\textcolor{blue}{\text{\tiny AAAI'25}}}$$^\dag$ & 96.8 & \textbf{\textcolor{red}{99.0}} & \textbf{\textcolor{red}{99.4}} & 91.7 & 95.4 & 96.4 \\
SelaVPR$_{\textcolor{blue}{\text{\tiny ICLR'24}}}$$^\dag$ & 95.7 & 98.8 & 99.2 & 90.6 & 95.3 & 96.2 \\
SelaVPR++$_{\textcolor{blue}{\text{\tiny TPAMI'25}}}$$^\dag$ & 96.6 & \textbf{\textcolor{red}{99.0}} & 99.3 & \textbf{\textcolor{red}{92.5}} & 95.7 & 96.5 \\
\rowcolor{lightshade}
FoL++ (global) & 95.6 & 98.8 & 99.2 & 91.6 & 95.6 & 96.5 \\
\rowcolor{lightshade}
FoL++$^\dag$   & \textbf{\textcolor{red}{96.9}} & \textbf{\textcolor{red}{99.0}} & \textbf{\textcolor{red}{99.4}} & 92.3 & \textbf{\textcolor{red}{95.9}} & \textbf{\textcolor{red}{96.8}} \\
\bottomrule
\end{tabular}
\label{tab:Pitts250k_Eynsham_results}
\vspace{-0.5cm}
\end{table}

\subsection{Implementation Details}

We initialize the backbone (ViT-B or ViT-L) with pre-trained DINOv2 weights, where the token dimensions are $768$ for ViT-B and $1024$ for ViT-L. Consistent with SALAD \citep{salad} and SuperVLAD \citep{supervlad}, we freeze most of the pre-trained backbone and directly fine-tune its last four transformer layers. This straightforward strategy avoids introducing additional Parameter-Efficient Fine-Tuning (PEFT) modules~\citep{sage} or Adapters~\citep{selavpr++}, while keeping our remaining task-specific modules learnable. In the feature extraction stage, the number of clusters $M$ is set to $64$. In the re-ranking stage, the number of channels in the up-convolutional (up-conv) blocks are $256$ and $128$, with a convolution kernel size of $3\times3$, stride=$2$, and padding=$1$.
Following SuperPoint~ \citep{detone2018superpoint} and XFeat~ \citep{potje2024xfeat}, we generate the spatial reliability map at $H/8 \times W/8$ to balance spatial granularity and efficiency.
The ViT architecture supports variable input sizes, provided images can be partitioned into $14 \times 14$ patches.
The final image similarity is computed by fusing global and local scores with a hyperparameter $\gamma = 1000$.
To speed up training, we used $224 \times 224$ images but evaluated on $322 \times 322$ resolution. Training was conducted on a single A100 GPU, using GSV-Cities\footnote{Available at \href{https://github.com/amaralibey/gsv-cities}{https://github.com/amaralibey/gsv-cities}.} as the training dataset, a curated dataset containing images from 23 cities collected from Google Street View~\citep{gsv}. The batch size is $60$, with each batch containing $4$ images. The AdamW optimizer with a linear learning rate schedule was used, with a learning rate of $6e-5$ and a weight decay of $9.5e-9$. The training converged after $5$ epochs. To ensure experimental validity and optimize hyperparameters, we monitored recall performance on the MSLS-val set. We follow mainstream works to use $25$ meters as the threshold for a correct scene and report recall@k (k=1,5,10) as evaluation metrics \citep{selavpr++,salad}.

\subsection{Comparisons with state-of-the-art Methods}

\begin{table*}[ht]
\centering
\setlength{\tabcolsep}{2.5pt} % 缩短列间距
\footnotesize 
\renewcommand{\arraystretch}{1.1} 
\caption{Memory footprint, latency, and performance comparison on Pitts30k and MSLS-Val. Evaluations are unified on an NVIDIA A100 GPU for fair comparison. Sub-columns under each dataset represent Recall@1 and Recall@5, respectively.}
\begin{tabular}{@{}lccccc c|c c|c@{}}
\toprule
Method & Local Features Dim. & Memory Footprint (GB)$\downarrow$ & Latency (s)$\downarrow$ & Device & Test-Res & \multicolumn{2}{c}{Pitts30k-test} & \multicolumn{2}{c}{MSLS-val} \\
\midrule
R2Former $_{\textcolor{blue}{\text{\tiny CVPR'23}}}$  & $500\times(128+3)$ & 0.244 & 0.115 & A100 & $322 \times 322$ & 91.1 & 95.2 & 89.7 & 95.0 \\
SelaVPR $_{\textcolor{blue}{\text{\tiny ICLR'24}}}$   & $61^2\times128$ & 0.182 & 0.060 & A100 & $322 \times 322$ & 92.8 & 96.8 & 90.8 & 96.4 \\
EffoVPR $_{\textcolor{blue}{\text{\tiny ICLR'25}}}$   & $649\times1024$ & 0.254 & \textbf{\textcolor{blue}{0.035}} & A100 & $504 \times 504$ & \textbf{\textcolor{blue}{93.9}} & \textbf{\textcolor{red}{97.4}} & \textbf{\textcolor{blue}{92.8}} & \textbf{\textcolor{blue}{97.2}} \\
FoL & $3500\times128$  & \textbf{\textcolor{blue}{0.171}} & 0.052 & A100 & $322 \times 322$ & \textbf{\textcolor{blue}{93.9}} & \textbf{\textcolor{blue}{96.9}} & 90.1 & 95.7 \\
\rowcolor{lightshade}
FoL++  & $3500\times128$    & \textbf{\textcolor{red}{0.130}} & \textbf{\textcolor{red}{0.031}} & A100 & $322 \times 322$ & \textbf{\textcolor{red}{94.7}}    & \textbf{\textcolor{red}{97.4}}   & \textbf{\textcolor{red}{94.1}}  & \textbf{\textcolor{red}{97.8}} \\
\bottomrule
\end{tabular}
\label{tab:comparison_mem}
\vspace{-0.2cm}
\end{table*}

\begin{figure*}[t]
    \centering
    \includegraphics[width=0.95\linewidth]{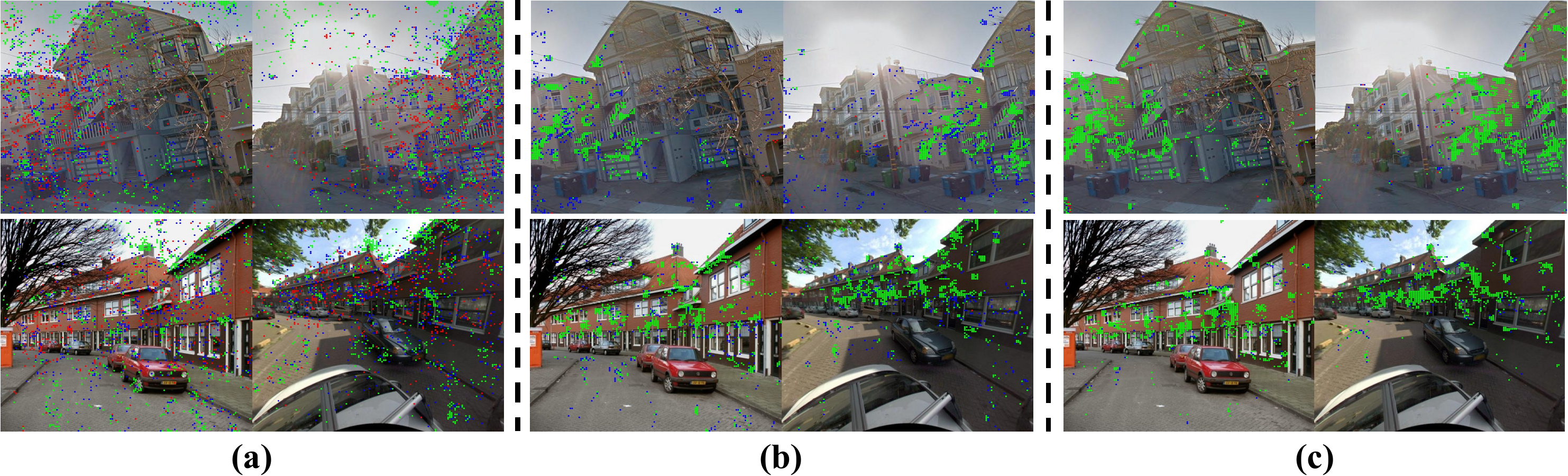}
    % \vspace{-0.2cm} % 这里调整图像与图注的间距，负值缩小间距
    \caption{Visualization of local feature correspondences in the re-ranking stage. (a) represents results \textit{w/o} Discriminative Region Guidance (DRG), while (b) shows results of \textbf{FoL} \textit{w/} DRG. (c) presents the results of \textbf{FoL++}, which demonstrates superior feature matching performance compared to \textbf{FoL}. Red $\color{red}\bullet$, Blue $\color{blue}\bullet$, and Green $\color{green}\bullet$ denote low, medium, and high similarity matching points, respectively, where similarity increases from ${\color{red}\bullet} \rightarrow {\color{blue}\bullet} \rightarrow {\color{green}\bullet}$.}
    \label{fig:Matching_demo}
    % \vspace{-0.3cm}
\end{figure*}

In this section, we compare the one-stage and two-stage results of our proposed FoL++ method with previous SoTA VPR methods. These methods include various one-stage approaches that utilize global feature retrieval: NetVLAD \citep{netvlad}, CosPlace \citep{cosplace}, MixVPR \citep{mixvpr}, EigenPlaces \citep{eigenplaces}, SALAD \citep{salad}, SALAD-CM \citep{salad-cm}, BoQ \citep{boq}, SuperVLAD \citep{supervlad}, EDTFormer \citep{edtformer}, MegaLoc \citep{megaloc}, ImAge \citep{image}, and SAGE \citep{sage}. Additionally, we compare our method with two-stage approaches that incorporate re-ranking: Patch-NetVLAD \citep{patchvlad}, TransVPR \citep{transvpr}, SelaVPR \citep{selavpr}, and SelaVPR++ \citep{selavpr++}.
Recent studies (i.e., SALAD, BoQ, SelaVPR++) as well as our FoL++ all employ the DINOv2 backbone for feature extraction, and different techniques are adopted to fine-tune the backbone for improved performance. In addition, SelaVPR performs fine-tuning on MSLS and further fine-tunes on Pitts30k, while EigenPlaces uses a large-scale dataset (SF-XL) constructed by CosPlace for training. Similar to SALAD, ImAge and SAGE, our method achieves highly competitive results.

Table~\ref{tab:compare_SOTA} presents the quantitative comparisons. Even at its global stage, our FoL++ method achieves highly competitive results against recent state-of-the-art single-stage methods like MegaLoc and SAGE. Notably, FoL++ (global) demonstrates a substantial improvement over the original FoL (global), particularly on the challenging Nordland and Nordland$\star$ datasets where the absolute R@1 jumps from 83.8\% to 91.8\% and 74.1\% to 84.5\%, respectively. On the Pitts30k dataset, FoL++ (global) surpasses top-performing methods like SelaVPR and BoQ. This strong performance is achieved with a compact feature dimension of 8,448 compared to BoQ's 16,384, underscoring the effectiveness of our learned global representation.

After re-ranking with local features, the complete FoL++ method demonstrates exceptional performance. Importantly, our method exhibits significant gains over the baseline FoL in the re-ranking stage as well. For instance, FoL++ improves the absolute R@1 by 0.8\%, 1.0\%, and 8.2\% compared to FoL on Pitts30k, Tokyo24/7, and Nordland, respectively. These results demonstrate our model's robustness against diverse challenges, including variations in illumination (Tokyo24/7), seasons (Nordland), viewpoints (Pitts30k), and dynamic conditions (MSLS).

Subsequently, we further evaluate our method on two general-purpose datasets: Pitts250k, which is suitable for large-scale urban evaluations, and Eynsham, a challenging grayscale dataset. As shown in Table~\ref{tab:Pitts250k_Eynsham_results}, when utilizing the ViT-L architecture, FoL++ achieves new state-of-the-art results across almost all metrics on both datasets. When specifically compared to the highly competitive SelaVPR++, FoL++ demonstrates superior performance in most cases, though SelaVPR++ retains a slight edge in the R@1 metric on Eynsham (92.5\% vs 92.3\%). Nevertheless, FoL++ still attains the highest R@5 and R@10 scores on Eynsham (reaching an impressive 96.8\% for R@10) and completely outperforms all baselines on the large-scale Pitts250k. This consistent excellence in both the global and re-ranking stages underscores its superior localization capability in various complex scenarios.

\begin{table}[h!]
\centering
\vspace{-0.3cm}
\caption{Performance and efficiency comparison with SOTA image matching methods\protect\footnotemark. Methods are evaluated at an inference resolution of $322 \times 322$. Metrics are measured on a single NVIDIA RTX A6000 GPU. For the image matching baselines, re-ranking is performed by sorting candidates based on RANSAC inliers.}
\label{tab:image_matching_comparison}

\setlength{\tabcolsep}{4pt} 
\fontsize{7.8pt}{9.0pt}\selectfont
\renewcommand{\arraystretch}{1.0} 

\begin{tabular}{l cccc}
\toprule
\multirow{2}{*}{Method} & \multicolumn{3}{c}{MSLS-val} & Matching \\
\cmidrule(lr){2-4}
& R@1 & R@5 & R@10 & Time (s) \\
\midrule
\rowcolor{lightblue}
\multicolumn{5}{l}{\textit{Image Matching Methods}} \\
RDD+MNN $_{\textcolor{blue}{\text{\tiny CVPR'25}}}$ & \textbf{\textcolor{blue}{89.3}} & 93.2 & 95.3 & 10.90 \\
xFeat+MNN $_{\textcolor{blue}{\text{\tiny CVPR'24}}}$ & 86.6 & 92.7 & 94.4 & \textbf{\textcolor{red}{2.788}} \\
SuperPoint+LG & 85.4 & 93.9 & 95.5 & 4.209 \\
RDD+LG & 88.9 & 93.9 & 95.4 & 12.57 \\
LoFTR $_{\textcolor{blue}{\text{\tiny CVPR'21}}}$ & 88.7 & \textbf{\textcolor{blue}{94.1}} & \textbf{\textcolor{blue}{95.9}} & 4.946 \\
E-LoFTR $_{\textcolor{blue}{\text{\tiny CVPR'24}}}$ & \textbf{\textcolor{red}{89.5}} & \textbf{\textcolor{red}{94.4}} & \textbf{\textcolor{red}{96.2}} & \textbf{\textcolor{blue}{3.939}} \\
\midrule
\rowcolor{lightblue}
\multicolumn{5}{l}{\textit{Two-Stage VPR Methods}} \\
SelaVPR $_{\textcolor{blue}{\text{\tiny ICLR'24}}}$ & \textbf{\textcolor{blue}{90.8}} & \textbf{\textcolor{blue}{96.4}} & \textbf{\textcolor{blue}{97.2}} & 0.085 \\
FoL & 90.1 & 95.7 & 96.9 & \textbf{\textcolor{blue}{0.078}} \\
\rowcolor{lightshade}
FoL++ & \textbf{\textcolor{red}{94.1}} & \textbf{\textcolor{red}{97.8}} & \textbf{\textcolor{red}{98.1}} & \textbf{\textcolor{red}{0.047}} \\
\bottomrule
\end{tabular}
\vspace{-0.3cm}
\end{table}
\footnotetext{Evaluation code: \url{https://github.com/gmberton/vismatch}.}

\begin{table*}[!htbp]
    \centering
    \setlength{\tabcolsep}{0.3mm} % 调整列间距
    \scriptsize
    \renewcommand{\arraystretch}{1} % 调整行间距
    \caption{Evaluation results of FoL, SelaVPR++, and FoL++ on five datasets. All methods use the ViT-B backbone and are evaluated at $322 \times 322$ resolution.}
    \label{tab:combined_results}
    \begin{tabular}{l c c ccc ccc ccc ccc ccc c}
    \toprule
    \multicolumn{3}{c}{Setting} & \multicolumn{3}{c}{Pitts30k-test} & \multicolumn{3}{c}{MSLS-val} & \multicolumn{3}{c}{Nordland} & \multicolumn{3}{c}{Tokyo24/7} & \multicolumn{3}{c}{Nordland$\star$} & \multicolumn{1}{c}{Avg.} \\
    \cmidrule(lr){1-3} \cmidrule(lr){4-6} \cmidrule(lr){7-9} \cmidrule(lr){10-12} \cmidrule(lr){13-15} \cmidrule(lr){16-18} \cmidrule(lr){19-19}
    Type & Test-Res & Stage & R@1 & R@5 & R@10 & R@1 & R@5 & R@10 & R@1 & R@5 & R@10 & R@1 & R@5 & R@10 & R@1 & R@5 & R@10 & R@1 \\
    \midrule
    \rowcolor{lightblue}
    \multicolumn{19}{c}{FoL} \\
    \midrule
    \multirow{2}{*}{ViT-B} & \multirow{2}{*}{$322 \times 322$} & Global & 92.1 & 96.4 & 97.6 & 91.1 & 95.7 & 96.4 & 72.7 & 85.5 & 89.6 & 94.6 & 96.5 & 96.8 & 62.5 & 80.3 & 85.0 & 82.6 \\
                           &                                   & Rerank & 93.1 & \textbf{\textcolor{red}{96.9}} & 97.7 & 91.5 & 96.2 & 96.8 & 85.4 & 92.7 & 94.8 & 97.5 & 98.1 & 98.4 & 78.2 & 90.2 & 92.9 & 89.1 \\
    \midrule
    \rowcolor{lightblue}
    \multicolumn{19}{c}{SelaVPR++ $_{\textcolor{blue}{\text{\tiny TPAMI'25}}}$} \\
    \midrule
    \multirow{2}{*}{ViT-B} & \multirow{2}{*}{$322 \times 322$} & Global & 89.6 & 95.2 & 96.7 & 89.1 & 95.7 & 96.6 & 79.0 & 91.0 & 94.3 & 89.8 & 96.5 & 97.5 & 67.5 & 85.7 & 90.3 & 83.0 \\
                           &                                   & Rerank & 93.2 & 96.6 & 97.6 & \textbf{\textcolor{red}{94.3}} & \textbf{\textcolor{red}{97.2}} & 97.3 & 94.9 & 97.8 & 98.4 & 97.5 & \textbf{\textcolor{red}{99.0}} & \textbf{\textcolor{red}{99.0}} & 87.0 & 95.3 & 96.6 & 93.4 \\
    \midrule
    \rowcolor{lightshade}
    \multicolumn{19}{c}{FoL++ (Ours)} \\
    \midrule
    \multirow{2}{*}{ViT-B} & \multirow{2}{*}{$322 \times 322$} & Global & 92.2 & 96.5 & 97.6 & 93.2 & 96.6 & 97.4 & 90.2 & 96.2 & 97.6 & 95.6 & 97.5 & 98.1 & 83.3 & 93.5 & 95.9 & 90.9 \\
                           &                                   & Rerank & \textbf{\textcolor{red}{93.3}} & \textbf{\textcolor{red}{96.9}} & \textbf{\textcolor{red}{97.9}} & 93.4 & 97.0 & \textbf{\textcolor{red}{97.6}} & \textbf{\textcolor{red}{95.1}} & \textbf{\textcolor{red}{98.2}} & \textbf{\textcolor{red}{98.8}} & \textbf{\textcolor{red}{97.8}} & 98.1 & 98.4 & \textbf{\textcolor{red}{89.1}} & \textbf{\textcolor{red}{96.5}} & \textbf{\textcolor{red}{97.6}} & \textbf{\textcolor{red}{93.7}} \\
    \bottomrule
    \end{tabular}
    % \vspace{-0.3cm}
\end{table*}

\begin{figure}[htp]
    \centering
    \includegraphics[width=0.95\linewidth]{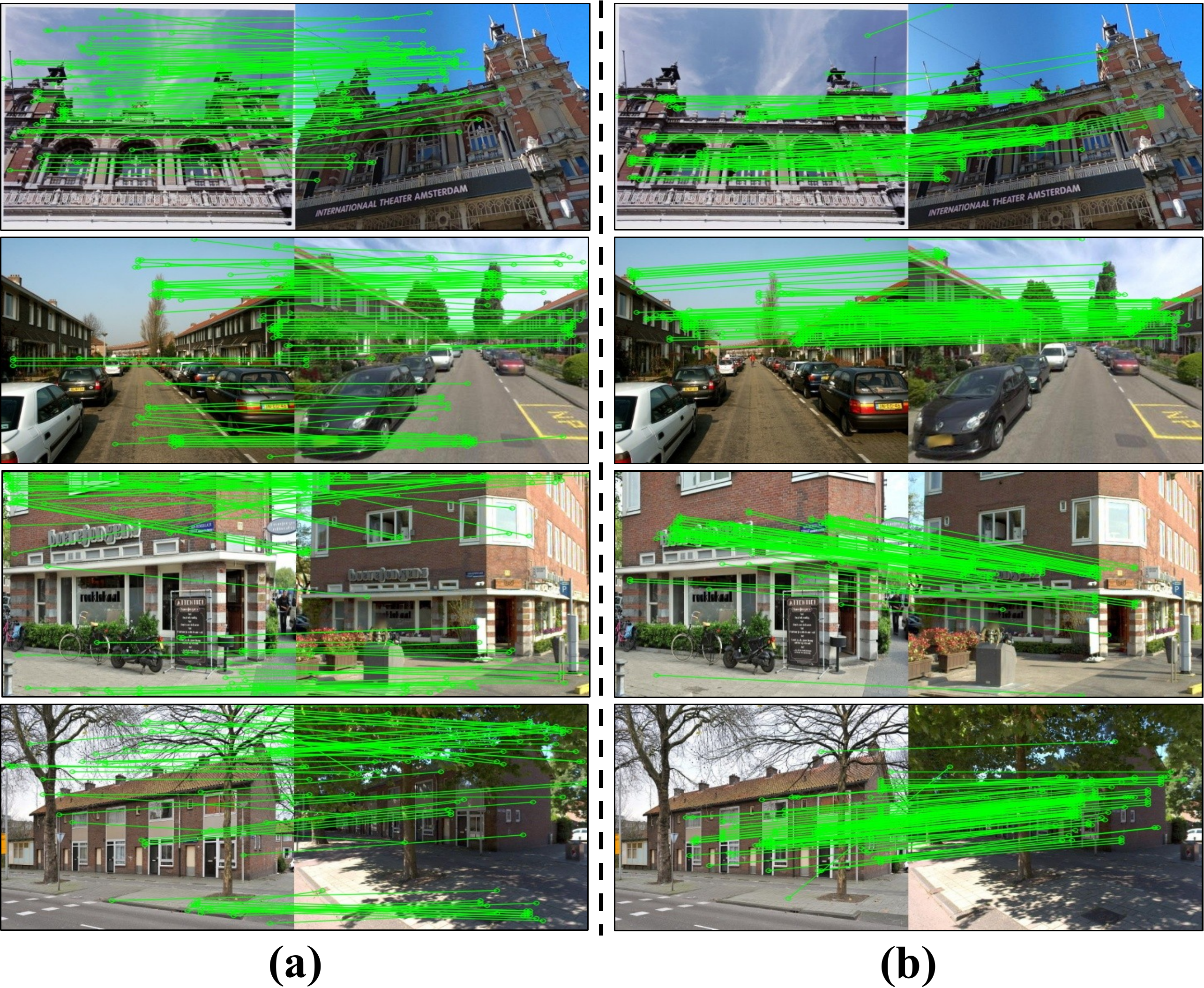}
        % \vspace{-0.3cm}
    \caption{Visualization of local feature matching in the re-ranking stage. (a) shows the w/o Discriminative Region Guidance, while (b) displays w/ Discriminative Region Guidance.
}
    \label{fig:FoL++_match}
    \vspace{-0.2cm}
\end{figure}

Efficiency is a crucial aspect when evaluating VPR methods. Table \ref{tab:comparison_mem} compares the computational complexity and memory footprint of various approaches. Notably, FoL++ achieves the lowest memory footprint (0.130GB) and the fastest runtime (0.031s) among all evaluated methods. Compared to our previous baseline, FoL, FoL++ further reduces the memory footprint from 0.171GB to 0.130GB and decreases the latency from 0.052s to 0.031s.
Furthermore, while EffoVPR demonstrates highly competitive latency (0.035s), it requires nearly double the memory (0.254GB). Compared to SelaVPR (0.182GB, 0.060s), FoL++ is significantly faster and more memory-efficient. By exclusively matching discriminative regions, FoL++ significantly reduces redundant computations while further improving retrieval accuracy.

Figure~\ref{fig:Matching_demo} illustrates the re-ranking stage via visual analysis of matching points and their confidence levels. The results indicate that FoL++ generates a higher concentration of high-confidence matches (green dots) while markedly reducing matches from less informative regions, such as the sky and ground, compared to FoL. This selective suppression of low-quality correspondences enhances both the robustness and precision of feature matching.
Figure~\ref{fig:FoL++_match} compares the effect of discriminative region guidance on local feature matching. Without this mechanism (Figure (a)), the matching process is overwhelmed by irrelevant regions, yielding numerous low-confidence matches that degrade retrieval precision. In contrast, with discriminative region guidance (Figure (b)), FoL++ selectively focuses on key features, such as structured architectural details, textured surfaces, or unique landmarks, thereby reducing mismatches and enhancing alignment reliability.

\begin{figure}[t!]
\setlength{\abovecaptionskip}{-0.02cm}
    \centering
    \includegraphics[width=1\linewidth]{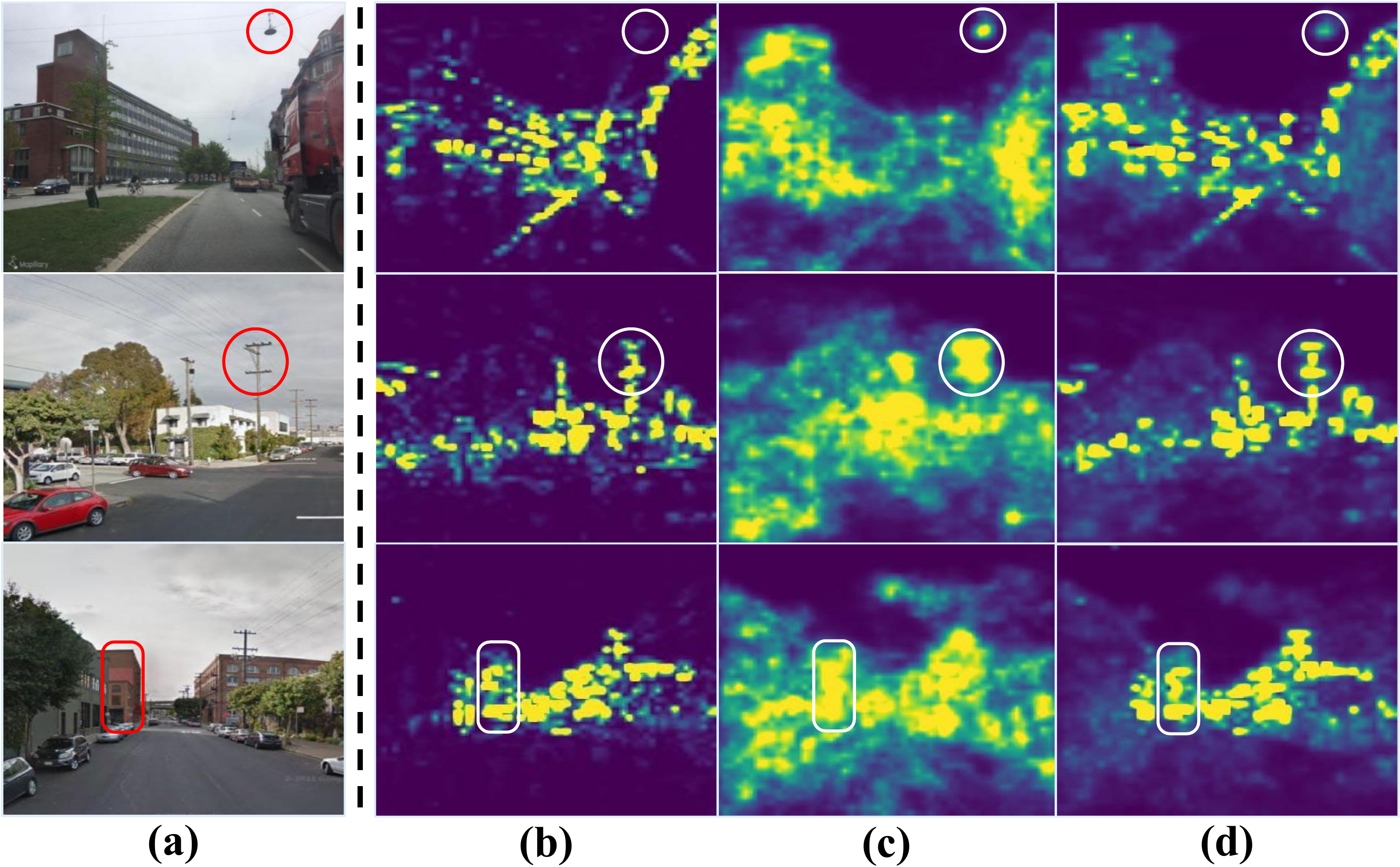}
\caption{Reliability map and discriminative mask. (a) Input scene. (b) Sinkhorn-based salient region mask. (c) REB's reliability heatmap. (d) Final discriminative mask, guiding local matching toward stable regions.}
\label{fig:FoL_plus_Map}
\vspace{-0.4cm}
\end{figure}

\begin{figure*}[!htbp]
    \centering
    \includegraphics[width=1\linewidth]{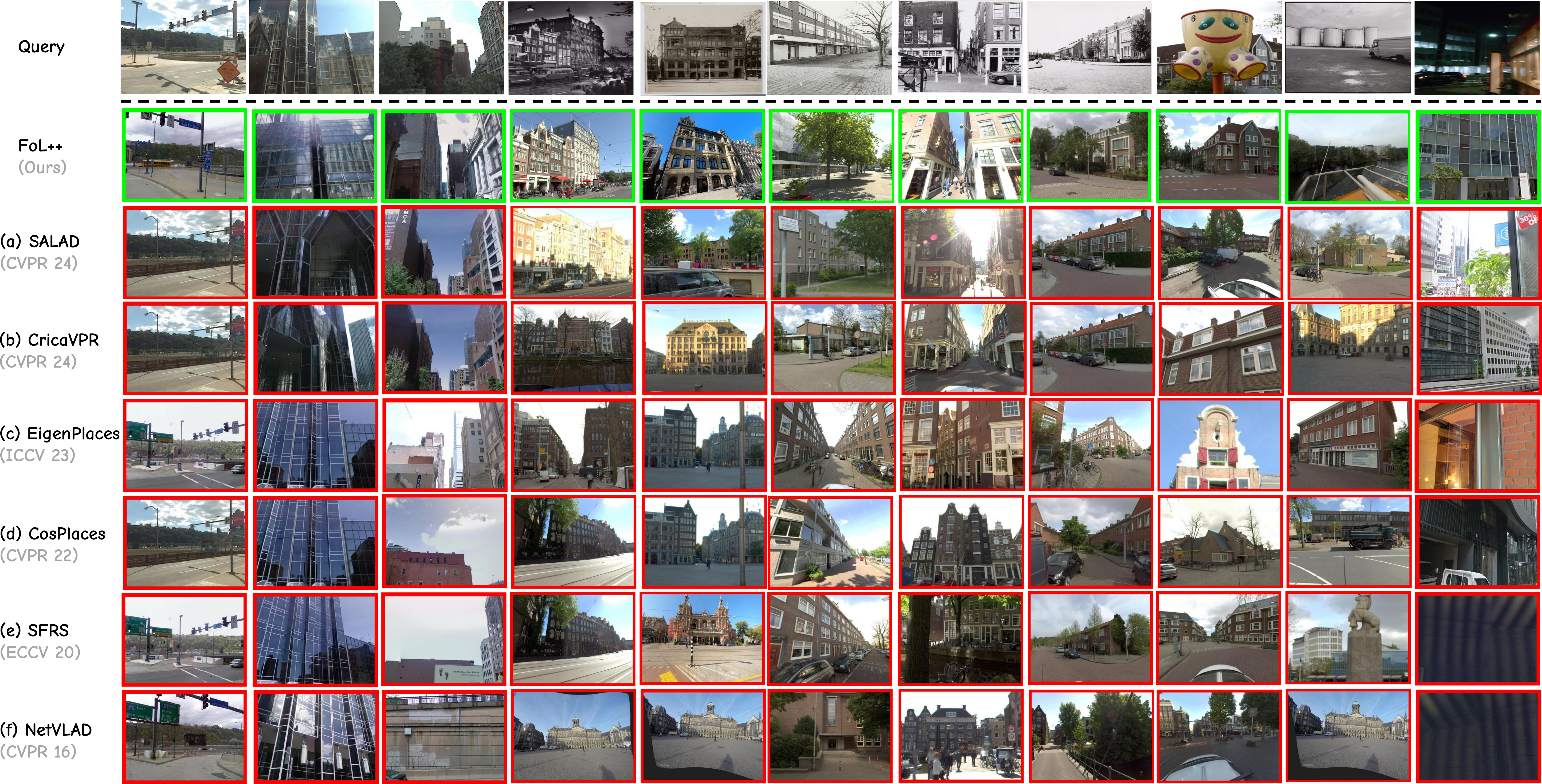}
    \caption{
        Qualitative Results. FoL++ achieves consistent accuracy in VPR across extreme viewpoint changes, illumination variations, and historical grayscale images. Correct matches (green-bordered) outperform methods like SALAD and CricaVPR, whose incorrect matches are highlighted in red, demonstrating robustness under complex real-world conditions. \textbf{Note:} The visualization is generated by adapting the evaluation toolkit from \url{https://github.com/gmberton/VPR-methods-evaluation}.
    }
    \label{fig:resultfig}
    \vspace{-0.1cm}
\end{figure*}

Figure~\ref{fig:FoL_plus_Map} visualizes the discriminative region modeling process, illustrating the difference between our proposed FoL++ and the original FoL framework across challenging scenarios, such as suspended streetlights, pole tops, and texture-less building segments (highlighted in red). As shown in column (b), the standard Sinkhorn-based mask, which relies on the baseline FoL approach, fails to capture these fine-grained or occlusion-prone regions. In contrast, the spatial reliability heatmap generated by the novel REB in FoL++ (c) explicitly models occlusion resilience, successfully highlighting these stable features. This mechanism produces the final discriminative mask (d) for FoL++, effectively guiding local matching toward reliable regions and improving localization performance under extreme viewpoint variations.
\begin{figure*}[!htbp]
\setlength{\abovecaptionskip}{-0.02cm}
    \centering
    \includegraphics[width=0.95\linewidth]{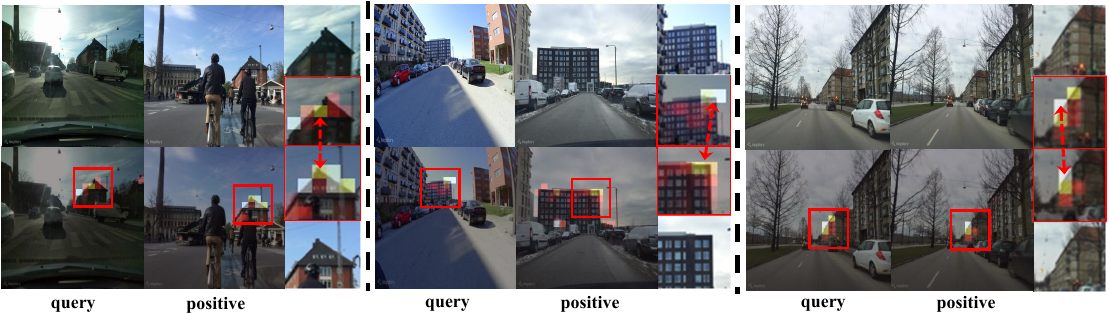}
    % \vspace{-0.6cm}
    \caption{
    The Visualization of Weakly Supervised Learning. Patches that are sufficiently discriminative and belong to the same cluster are used as weakly supervised training sample pairs.
    }
    % \vspace{-0.3cm}
    \label{fig:vis-ws}
\end{figure*}
As illustrated in Figure~\ref{fig:vis-ws}, during the global feature aggregation phase, local feature patches are assigned to clusters based on their discriminative properties. This clustering inherently encodes patch correspondence, which we utilize to construct pseudo-correspondence pairs for weakly supervised learning.
Figure~\ref{fig:resultfig} presents a qualitative comparison between FoL++ and SoTA methods under extreme conditions. Whether facing viewpoint shifts (e.g., top-down vs. street-level views), drastic illumination changes (e.g., historical black-and-white versus modern nighttime scenes), seasonal transitions, or heavy occlusions caused by crowds or moving vehicles, FoL++ consistently achieves accurate matches.

\begin{table}[t]
    \centering
    \setlength{\tabcolsep}{1.1mm}
    \caption{Ablation Study on Multiple Datasets. Evaluated at $322 \times 322$ resolution. Here DRG denotes the Efficient Re-ranking with Discriminative Region Guidance.}
    \vspace{-0.2cm}
    \small
    \renewcommand{\arraystretch}{1.1}
    \begin{tabular}{l|cc|cc|cc}
    \toprule
    \multirow{2}{*}{Method} & \multicolumn{2}{c|}{Pitts30k-test} & \multicolumn{2}{c|}{MSLS-val} & \multicolumn{2}{c}{Nordland} \\
    \cmidrule(lr){2-7}
     & R@1 & R@5 & R@1 & R@5 & R@1 & R@5 \\
    \hline
    Baseline              & 92.5  & 96.4  & 92.2  & 96.4  & 89.7  & 95.5 \\
    \hdashline
    $+ \text{re-ranking}$ & 93.3  & 96.8  & 92.9  & 96.8  & 91.5  & 96.5 \\
    $+ \mathcal{L}_{SA}$  & 93.7  & 97.1  & 93.3  & 97.0  & 92.5  & 97.0 \\
    $+ \mathcal{L}_{SCE}$  & 93.9  & 97.1  & 93.5  & 97.1  & 93.2  & 97.3 \\
    $+ \mathcal{L}_{PC}$  & 94.2  & 97.2  & 93.8  & 97.3  & 94.0  & 97.7 \\
    \hdashline
    \rowcolor{lightshade}
    $+ \text{DRG}$      & \textbf{\textcolor{red}{94.7}}  & \textbf{\textcolor{red}{97.4}}  & \textbf{\textcolor{red}{94.1}}  & \textbf{\textcolor{red}{97.8}}  & \textbf{\textcolor{red}{96.1}}  & \textbf{\textcolor{red}{98.6}} \\
    \bottomrule
    \end{tabular}
    \label{tab:ablation_loss}
    \vspace{-0.2cm}
\end{table}

\begin{table}[t]
    \centering
    \setlength{\tabcolsep}{0.6mm}
    \caption{Ablation study of Dynamic Top-$k$ Candidate Selection (DCS) vs. fixed $k$. Re-ranking time is measured in (s) and memory footprint in (GB). Both efficiency metrics report the average results evaluated on the Pitts30k-test dataset, tested on a single RTX 3090.}
    \vspace{-0.1cm}
    \renewcommand{\arraystretch}{1.1}
    \begin{tabular}{l|cc|cc|cc}
    \toprule
    \multirow{2}{*}{Method} & \multicolumn{2}{c|}{Pitts30k-test} & \multicolumn{2}{c|}{MSLS-val} & \multicolumn{2}{c}{Efficiency} \\
    \cmidrule(lr){2-7}
    & R@1 & R@5 & R@1 & R@5 & Re-rank$\downarrow$ & Mem$\downarrow$ \\
    \midrule
    Fixed k=10  & 93.6 & 96.6 & 93.1 & 96.9 & 0.63   & 1.30  \\
    Fixed k=30  & 94.0 & 96.9 & 93.4 & 97.2 & 1.90   & 3.90  \\
    Fixed k=50  & 94.3 & 97.1 & 93.8 & 97.5 & 3.16   & 6.50  \\
    Fixed k=75  & 94.4 & \textbf{\textcolor{blue}{97.3}} & \textbf{\textcolor{blue}{94.0}} & 97.5 & 4.74   & 9.75  \\
    Fixed k=100 & \textbf{\textcolor{blue}{94.5}} & \textbf{\textcolor{blue}{97.3}} & 93.9 & \textbf{\textcolor{blue}{97.6}} & 6.32   & 13.0  \\
    \midrule
    \rowcolor{lightshade}
    DCS (Ours)  & \textbf{\textcolor{red}{94.7}} & \textbf{\textcolor{red}{97.4}} & \textbf{\textcolor{red}{94.1}} & \textbf{\textcolor{red}{97.8}} & 3.59  & 7.41  \\
    \bottomrule
    \end{tabular}
    \label{tab:DCS_ablation}
    \vspace{-0.3cm}
\end{table}

\subsection{Ablation Study}

To comprehensively evaluate the performance of our proposed VPR re-ranking module, we compared FoL++ against state-of-the-art general-purpose image matching methods, including detector-based approaches (RDD~ \citep{chen2025rdd}, xFeat~ \citep{potje2024xfeat}, SuperPoint~ \citep{detone2018superpoint}) and detector-free matchers (LoFTR~ \citep{loftr}, Efficient-LoFTR~ \citep{wang2024efficient}). For these baselines, re-ranking is performed by strictly applying RANSAC to count the number of geometric inliers between the query and top-retrieved candidates.
As presented in Table~\ref{tab:image_matching_comparison}, FoL++ significantly outperforms all image matching baselines in both retrieval accuracy and computational efficiency. In terms of accuracy, while methods like Efficient-LoFTR achieve competitive results (89.5\% R@1 on MSLS-val), FoL++ surpasses this best-performing matcher by a clear margin of 4.6\%. This indicates that our discriminative region modeling is more effective for place recognition than generic geometric verification.
More importantly, FoL++ completes the matching process in only 0.047s, whereas the image matching baselines require anywhere from 2.788s to 12.57s. Furthermore, compared to previous specialized two-stage VPR methods like SelaVPR (0.085s) and our baseline FoL (0.078s), FoL++ achieves much higher accuracy while nearly halving the computational time. Coupled with its robust accuracy, this improved efficiency suggests that FoL++ holds promising potential for large-scale, real-time visual place recognition applications.

Table~\ref{tab:combined_results} details the performance of our FoL++, its predecessor FoL, and the recent state-of-the-art method SelaVPR++ across five datasets under a strictly controlled setting (ViT-B backbone evaluated at $322 \times 322$ resolution). 
At the global retrieval stage, FoL++ significantly outperforms both FoL and SelaVPR++. For instance, on the challenging Nordland$\star$ dataset, our global R@1 reaches 83.3\%, exceeding FoL (62.5\%) and SelaVPR++ (67.5\%) by a massive margin. This demonstrates that our region-aware training extracts much more robust global descriptors compared to existing methods.
After the re-ranking stage, FoL++ consistently shows superior top-1 localization accuracy across most benchmarks. It achieves the highest R@1 on Pitts30k-test (93.3\%), Nordland (95.1\%), Tokyo24/7 (97.8\%), and Nordland$\star$ (89.1\%), outperforming the re-ranking results of SelaVPR++. Although SelaVPR++ exhibits strong performance on the MSLS-val dataset (achieving 94.3\% R@1) and yields slightly higher R@5 and R@10 scores on Tokyo24/7, FoL++ remains highly competitive overall. By delivering the best R@1 performance on four out of five datasets, it highlights the effectiveness and broad generalization ability of our reliability-aware local matching mechanism.

\begin{figure}[!htbp]
    \centering
    \includegraphics[width=1\linewidth]{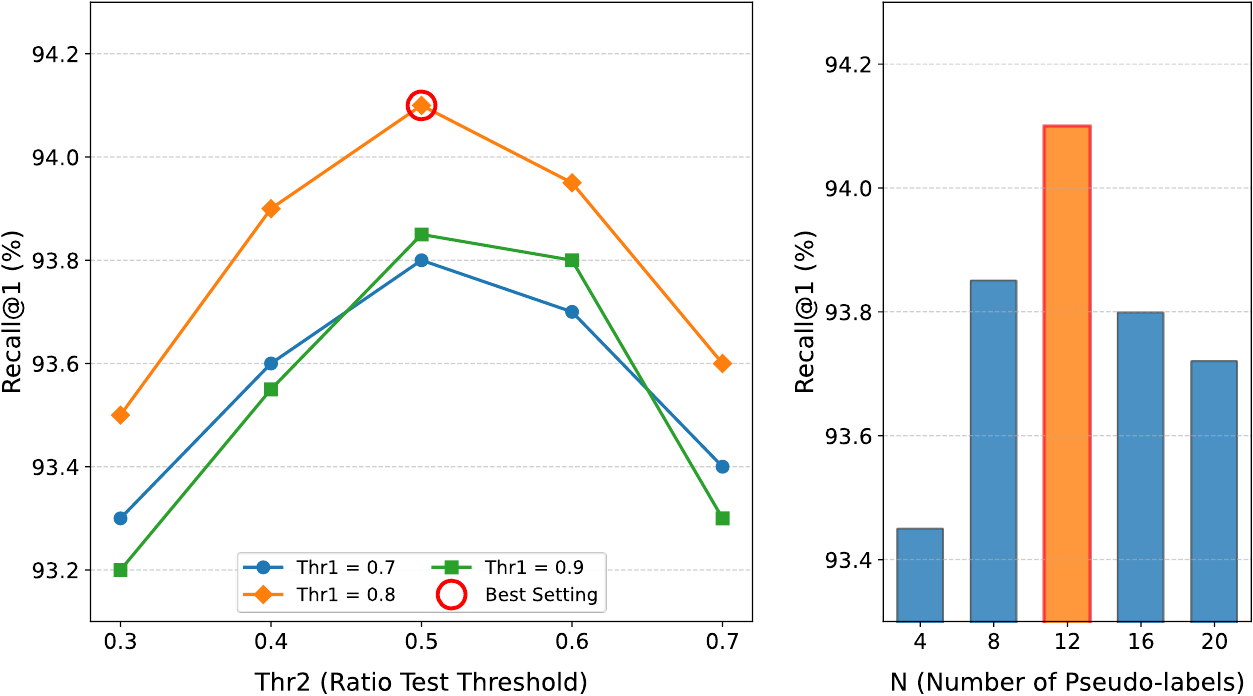}
    % \vspace{-0.6cm}
    \caption{
        Ablation Study of the Hyperparameters $thr1$, $thr2$ and $N$ on the MSLS-Val in the context of weakly supervised local feature learning.
    }
    % \vspace{-0.3cm}
    \label{fig:ws-ablation}
    \vspace{-0.2cm}
\end{figure}

Table~\ref{tab:ablation_loss} presents the ablation results across multiple datasets. The baseline uses the SALAD model trained solely with the $\mathcal{L}_{MS}$ loss. Applying re-ranking first effectively improves performance (e.g., R@1 on Pitts30k-test increases from 92.5\% to 93.3\%, and on Nordland from 89.7\% to 91.5\%). Adding $\mathcal{L}_{SA}$ for discriminative region modeling provides notable gains (+0.4\% on Pitts30k-test and +1.0\% on Nordland). $\mathcal{L}_{SCE}$ contributes further spatial refinements (+0.2\% on MSLS-val and +0.7\% on Nordland). $\mathcal{L}_{PC}$ continues to improve local feature representations (+0.3\% on both Pitts30k-test and MSLS-val). Finally, our proposed DRG (Efficient Re-ranking with Discriminative Region Guidance) achieves the best overall results. It notably yields a massive +2.1\% absolute improvement on the challenging Nordland dataset (reaching 96.1\% R@1) and establishes highly competitive performance on Pitts30k-test (94.7\% R@1), all while reducing matching time compared to dense local feature methods.

Table~\ref{tab:DCS_ablation} compares our Dynamic Top-$k$ Candidate Selection (DCS) method with fixed top-$k$ baselines. Unlike fixed $k$ strategies, where blindly increasing the candidate pool size yields diminishing returns or even introduces noisy candidates that degrade performance (e.g., R@1 on MSLS-val slightly drops at $k=100$), DCS dynamically filters out distracting candidates, achieving the highest accuracy across all metrics. Notably, while operating at an efficiency level highly comparable to the fixed $k=50$ baseline (with a re-ranking time of 3.59 s vs. 3.16 s), DCS significantly boosts the R@1 on Pitts30k-test from 94.3\% to 94.7\%. Furthermore, compared to the aggressive $k=100$ baseline, DCS reduces the re-ranking time by approximately 43\% (from 6.32 s to 3.59 s) and substantially decreases memory usage from 13.0 GB to 7.41 GB. These results confirm that the adaptive mechanism of DCS is not merely an efficiency optimization, but a crucial component to effectively suppress false positives, making it an optimal and scalable solution for large-scale image retrieval.

\begin{figure*}[!t]
    \centering
    \includegraphics[width=\linewidth]{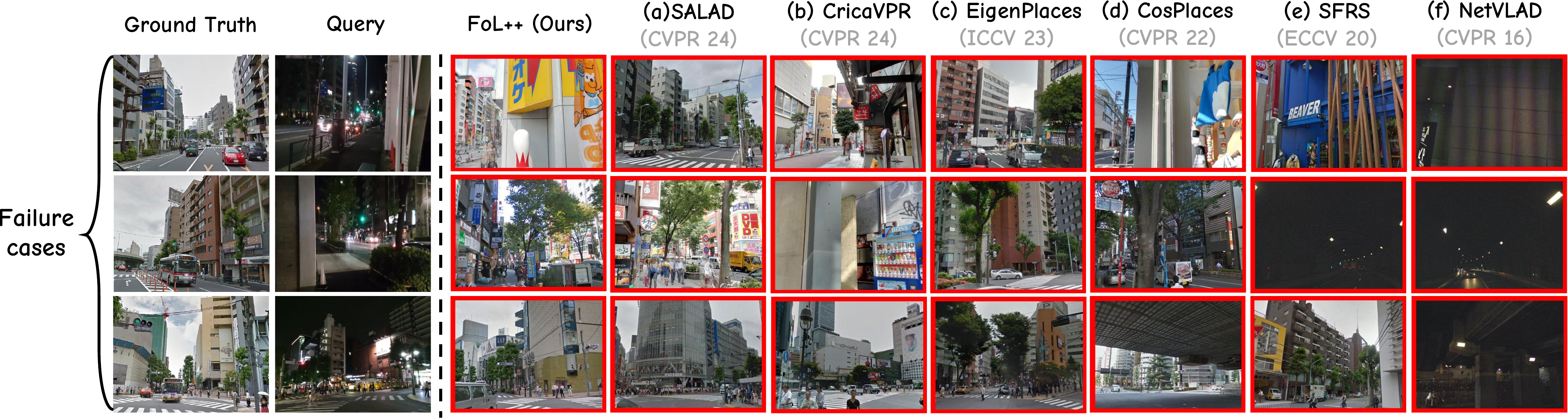}
    \caption{\textbf{Failure cases} caused by severe illumination changes, particularly day-to-night transitions. All evaluated methods, including ours, struggle under these extreme conditions.}
\label{fig:Failure_cases}
\vspace{-0.2cm}
\end{figure*}

Figure~\ref{fig:ws-ablation} presents the ablation study on the hyperparameters of our weakly supervised framework: similarity threshold $thr1$, ratio test threshold $thr2$, and pseudo-label count $N$. As illustrated in the left plot, the configuration of $thr1=0.8$ and $thr2=0.5$ achieves the highest Recall@1 (94.1\%) on the MSLS-val dataset. Deviating from these optimal values degrades performance: thresholds that are too low or loose introduce low-confidence noise and repetitive structures, whereas overly high or strict thresholds excessively discard valid labels and diverse features. Similarly, the right plot shows that the performance peaks at $N=12$. A smaller $N$ restricts the diversity of the training signal, while an excessively large $N$ degrades matching quality by introducing lower-quality candidates.

To quantitatively assess the reliability of our weakly supervised strategy, we evaluate the generated pseudo-labels against a proxy ground truth established by SuperPoint, LightGlue, and RANSAC. Since our pseudo-labels represent patch-level correspondences ($14 \times 14$ pixels at $322 \times 322$ resolution), we approximate their spatial locations using patch centers. To reasonably account for this inherent spatial discretization, a match is considered valid if its epipolar distance to the oracle's estimation is within 7 pixels (i.e., half the patch size). 
Under optimal hyperparameters ($thr1=0.8$, $thr2=0.5$, $N=12$), 84.3\% of the pseudo-labels from a random GSV-Cities subset satisfy this geometric constraint. Furthermore, a baseline using random intra-cluster matches yields a precision of only 62.5\% under the same 7-pixel tolerance, confirming that our strategy captures genuine geometric alignments rather than overlaps. Moreover, to handle residual noise, our loss function $\mathcal{L}_{PC}$ employs a soft-weighting mechanism via the $\exp(\text{sim}(f_{p_i}, f'_{p_i}))$ term, acting as a confidence gate to further suppress ambiguous matches.

\begin{table}[htbp]
\centering
\setlength{\tabcolsep}{1.2mm}
\caption{Ablation on the threshold percentage for binarizing the region mask \(\mathbf{M}\). We vary the top percentage \(k\) used to set elements to 1.}
\renewcommand{\arraystretch}{1.1}
\begin{tabular}{@{}c ccc ccc@{}}
\toprule
\multirow{2}{*}{Threshold (\%)} & \multicolumn{3}{c}{Pitts30k-test} & \multicolumn{3}{c}{MSLS-val} \\
\cmidrule(lr){2-4} \cmidrule(lr){5-7} 
                                & R@1  & R@5  & R@10  & R@1  & R@5  & R@10  \\ \midrule
20                              & 92.5  & 96.6  & 97.5  & 93.1  & 96.9  & 97.2  \\
30                              & \textcolor{blue}{\textbf{94.3}}  & 97.1  & 97.9  & \textcolor{blue}{\textbf{93.9}}  & \textcolor{blue}{\textbf{97.6}}  & \textcolor{blue}{\textbf{97.8}}  \\
\rowcolor{lightshade}
40                              & \textcolor{red}{\textbf{94.7}} & \textcolor{red}{\textbf{97.4}} & \textcolor{red}{\textbf{98.1}} & \textcolor{red}{\textbf{94.1}} & \textcolor{red}{\textbf{97.8}} & \textcolor{red}{\textbf{98.1}} \\
50                              & 94.2  & \textcolor{blue}{\textbf{97.2}}  & \textcolor{blue}{\textbf{98.0}}  & 93.5  & 97.3  & 97.7  \\
60                              & 93.8  & 96.9  & 97.9  & 93.8  & 97.4  & 97.4  \\ \bottomrule
\end{tabular}
\label{tab:thresh_ablation}
\vspace{-0.4cm}
\end{table}

\begin{table}
    \centering
    \setlength{\tabcolsep}{0.4mm}
    \caption{Ablation study of Reliability-Aware Local Matching (RALM) and Similarity Computation (SC).}
    \begin{tabular}{c c c c c c c c c}
    \toprule
    \multirow{2}{*}{Method} & \multicolumn{2}{c}{Our Strategies} & \multicolumn{3}{c}{Pitts30k-test} & \multicolumn{3}{c}{MSLS-val} \\
    \cmidrule(lr){2-3} \cmidrule(lr){4-6} \cmidrule(lr){7-9}
    & RALM & SC & R@1 & R@5 & R@10 & R@1 & R@5 & R@10 \\
    \midrule
    \multirow{4}{*}{FoL++}
    &  &  & 94.2 & 97.0 & 97.8 & 93.7 & 97.3 & 97.5 \\
    & \checkmark &  & 94.3 & 97.1 & \textcolor{blue}{\textbf{98.0}} & \textcolor{blue}{\textbf{93.9}} & \textcolor{blue}{\textbf{97.6}} & \textcolor{blue}{\textbf{97.9}} \\
    &  & \checkmark & \textcolor{blue}{\textbf{94.5}} & \textcolor{blue}{\textbf{97.2}} & \textcolor{blue}{\textbf{98.0}} & 93.8 & 97.4 & 97.7 \\
    & \checkmark & \checkmark & \cellcolor{lightshade}\textcolor{red}{\textbf{94.7}} & \cellcolor{lightshade}\textcolor{red}{\textbf{97.4}} & \cellcolor{lightshade}\textcolor{red}{\textbf{98.1}} & \cellcolor{lightshade}\textcolor{red}{\textbf{94.1}} &\cellcolor{lightshade}\textcolor{red}{\textbf{97.8}} & \cellcolor{lightshade}\textcolor{red}{\textbf{98.1}} \\
    \bottomrule
    \end{tabular}
\label{tab:ralm_sc_ablation}
\vspace{-0.4cm}
\end{table}

Table~\ref{tab:thresh_ablation} compares the performance using different thresholds for binarizing the region mask \(\mathbf{M}\), where the top \(k\%\) elements are set to 1 and the others to 0. Among the evaluated configurations, the 40\% threshold (highlighted in red) yields the highest scores across all metrics on both datasets. Specifically, setting a lower threshold (e.g., 20\%) results in a drop in accuracy (e.g., yielding 92.5\% R@1 on Pitts30k-test), which can be attributed to the loss of critical spatial information. On the other hand, a higher threshold (e.g., 60\%) also degrades the matching precision, due to the introduction of irrelevant background noise. Based on these empirical results, we adopt the 40\% threshold for region binarization in our framework.

Table~\ref{tab:ralm_sc_ablation} shows that both Reliability-Aware Local Matching (RALM) and Similarity Computation (SC) can independently enhance retrieval performance over the baseline, offering complementary benefits across different scenarios. For instance, on the challenging MSLS-val dataset, RALM plays a more prominent role, boosting the R@1 from 93.7\% to 93.9\%. On the other hand, on the Pitts30k-test dataset, SC provides a more substantial gain, improving the R@1 from 94.2\% to 94.5\%. Ultimately, the strongest performance is achieved by combining both RALM and SC. The full model attains 94.7\% and 94.1\% R@1 on Pitts30k-test and MSLS-val respectively, effectively surpassing all individual configurations.

\section{Discussion}
\label{sec:discussion}

To address the flaws in traditional two-stage VPR pipelines, FoL++ introduces the Reliability Estimation Branch to explicitly model occlusion resistance. Building on this, our Reliability-Aware Local Matching mechanism dynamically weights local correspondences, penalizing matches in unstable regions while highlighting reliable landmarks. Furthermore, the Adaptive Candidate Scheduler overcomes the computational bottlenecks of static re-ranking. This unified paradigm effectively bridges the gap between coarse global filtering and fine-grained local matching, establishing a scalable balance between accuracy and speed.

Despite these systematic advancements, both our method and the broader VPR paradigm face inherent challenges. First, as performance on standard benchmarks nears saturation, controversial and incorrect labels inherent in the datasets also exert a impact. Current evaluations depend on the task-defined geographic threshold (25 meters), which often ignores visual overlap and exacerbates this label ambiguity. Second, under extreme illumination variations, relying on a single foundation model's architecture can still limit representation capacity when photometric cues vanish. 

To address these limitations and further advance the field, we suggest exploring the following four directions for future research:
\begin{itemize}
    \item \textbf{Feature Complementarity \& Architectural Evolution:} To combat benchmark saturation and illumination challenges, we suggest that future models could explore feature mixing from the latest vision architectures \citep{dinov3,SAM3}, leveraging complementary representations to maintain robust spatial saliency across varying conditions.
    \item \textbf{Pose-Aware Two-Stage Refinement:} To focus on the geographic threshold, future frameworks could incorporate lightweight pose estimation during the local matching stage \citep{reloc3r}. This would provide metric-aware geometric verification and handle distance threshold information more dynamically.
    \item \textbf{Efficiency Scaling \& Representation Refinement:} To further push the boundaries of deployment efficiency, techniques such as multi-teacher knowledge distillation and learnable linear projections for dimensionality reduction offer potential \citep{cricavpr}. Additionally, using fine-grained local features to iteratively refine global descriptors could yield more expressive representations \citep{embodiedplace}.
    \item \textbf{Data Generation \& Novel Modalities:} To overcome dataset biases, diffusion models can be utilized to construct pose-controllable, multi-view training sets. Furthermore, integrating interactive vision language models, or introducing event streams for high dynamic-range scenarios, could expand VPR from a retrieval task into a context-aware navigational dialogue.
\end{itemize}

\section{Conclusion}
\label{sec:conclusion}

FoL++ provides an effective framework to address key challenges in VPR by introducing a reliability-aware discriminative region learning paradigm. Specifically, we propose the Reliability Estimation Branch (REB) to explicitly model occlusion resistance via spatial reliability maps. This representation is efficiently optimized by two spatial alignment losses (SAL and SCEL) to highlight discriminative regions and mitigate perceptual aliasing. Additionally, the designed pseudo-correspondence strategy enables weakly supervised local feature training without pixel-level annotations by leveraging global feature cluster consistency. Our Adaptive Candidate Scheduler dynamically adapts candidate pools and weights local matches via global similarity and spatial reliability, surpassing traditional independent systems. Extensive experiments across seven benchmarks demonstrate that FoL++ achieves SoTA performance with 94.7\% R@1 on Pitts30k-test and 94.1\% on MSLS-val. It maintains a memory footprint of only 0.13 GB and achieves 40\% computational savings compared to FoL. The proposed cross-stage fusion mechanism effectively bridges global and local visual evidence, delivering robust localization under extreme viewpoint changes, illumination variations, and occlusions, thereby establishing a scalable and reliable foundation for efficient VPR in complex urban environments.

\bibliography{sn-bibliography}% common bib file

\end{document}